\documentclass[10pt,twocolumn,letterpaper]{article}

\usepackage[pagenumbers]{cvpr} %

\usepackage{graphicx}
\usepackage{amsmath}
\usepackage{amssymb}
\usepackage{booktabs}
\usepackage{graphicx}
\usepackage{fancyhdr}
\usepackage{float}
\usepackage{xcolor}
\usepackage[pagebackref=true,breaklinks=true,letterpaper=true,colorlinks,citecolor=citecolor,bookmarks=false]{hyperref}
\usepackage{textpos}
\usepackage[table]{colortbl}
\definecolor{citecolor}{RGB}{30,102,235}

\usepackage[capitalize]{cleveref}
\crefname{section}{Sec.}{Secs.}
\Crefname{section}{Section}{Sections}
\Crefname{table}{Table}{Tables}
\crefname{table}{Tab.}{Tabs.}
\def\etal{\emph{et al.}\@\xspace}
\def\cvprPaperID{***} %
\def\confName{CVPR}
\def\confYear{2023}
\begin{document}

\newcommand{\sx}[1]{\textcolor{cyan}{sx: #1}}
\newcommand{\bp}[1]{\textcolor{cyan}{bp: #1}}
\definecolor{deemph}{gray}{0.6}
\newcommand{\gb}{\rowcolor{gray!20}}

\newcommand\blfootnote[1]{\begingroup\renewcommand\thefootnote{}\footnote{#1}\addtocounter{footnote}{-1}\endgroup}

\title{Scalable Diffusion Models with Transformers}

\author{William Peebles\textsuperscript{*}\\
UC Berkeley
\and
Saining Xie\\
New York University
}

\twocolumn[{%
\renewcommand\twocolumn[1][]{#1}%
\maketitle
\vspace{-8mm}
\begin{center}
    \centering
    \captionsetup{type=figure}
    \includegraphics[width=\linewidth]{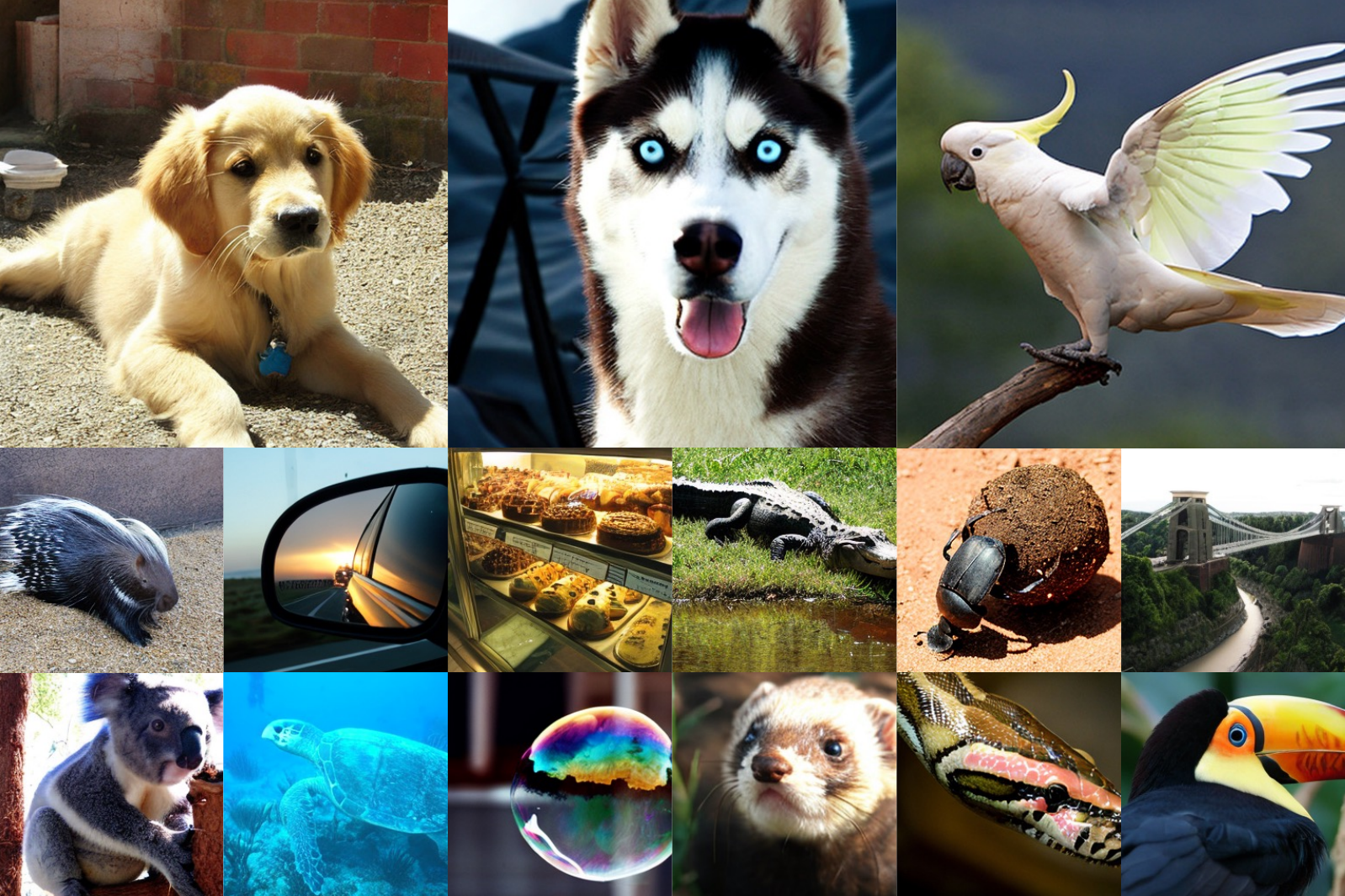}
    \vspace{-4mm}
    \captionof{figure}{\textbf{Diffusion models with transformer backbones achieve state-of-the-art image quality.} We show selected samples from two of our class-conditional DiT-XL/2 models trained on ImageNet at 512$\times$512 and 256$\times$256 resolution, respectively.}\label{fig:teaser}
\end{center}%
}]
\begin{abstract}
\vspace{-2mm}
We explore a new class of diffusion models based on the transformer architecture. We train latent diffusion models of images, replacing the commonly-used U-Net backbone with a transformer that operates on latent patches. We analyze the scalability of our Diffusion Transformers (DiTs) through the lens of forward pass complexity as measured by Gflops. We find that DiTs with higher Gflops---through increased transformer depth/width or increased number of input tokens---consistently have lower FID. In addition to possessing good scalability properties, our largest DiT-XL/2 models outperform all prior diffusion models on the class-conditional ImageNet 512$\times$512 and 256$\times$256 benchmarks, achieving a state-of-the-art FID of 2.27 on the latter.
\vspace{10mm}
\end{abstract}

\vspace{-8.4mm}
\section{Introduction}

\blfootnote{\textsuperscript{*} Work done during an internship at Meta AI, FAIR Team.}
\blfootnote{\ \ \ Code and project page available \href{https://www.wpeebles.com/DiT.html}{here}.}

\begin{figure*}[t]
\includegraphics[width=1.0\linewidth]{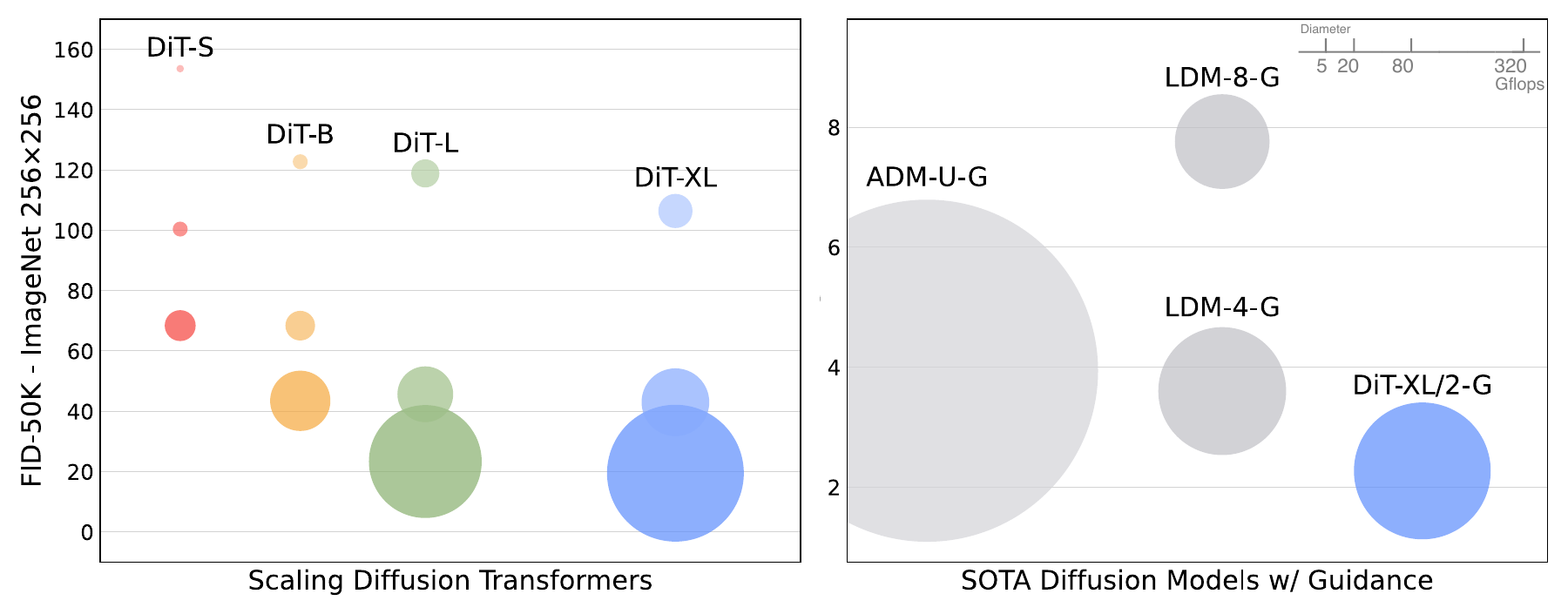}
\vspace{-4mm}
\caption{\textbf{ImageNet generation with Diffusion Transformers (DiTs).} Bubble area indicates the flops of the diffusion model. \emph{Left:} FID-50K (lower is better) of our DiT models at 400K training iterations. Performance steadily improves in FID as model flops increase. \emph{Right:} Our best model, DiT-XL/2, is compute-efficient and outperforms all prior U-Net-based diffusion models, like ADM and LDM.}
\label{fig:bubbles}
\end{figure*}
\vspace{-4mm}
Machine learning is experiencing a renaissance powered by transformers. Over the past five years, neural architectures for natural language processing~\cite{Radford2018,Devlin2019}, vision~\cite{Dosovitskiy2020} and several other domains have largely been subsumed by transformers~\cite{Vaswani2017}. 
Many classes of image-level generative models remain holdouts to the trend, though---while transformers see widespread use in autoregressive models~\cite{Radford2019,Brown2020,Chen2020generative,ramesh2021zero}, they have seen less adoption in other generative modeling frameworks. 
For example, diffusion models have been at the forefront of recent advances in image-level generative models~\cite{dhariwal2021adm,ramesh2022hierarchical}; yet, they all adopt a convolutional U-Net architecture as the de-facto choice of backbone. 

The seminal work of Ho \etal~\cite{ho2020ddpm} first introduced the U-Net backbone for diffusion models. Having initially seen success within pixel-level autoregressive models and conditional GANs~\cite{isola2017image}, the U-Net was inherited from PixelCNN++~\cite{salimans2017pixelcnn++,van2016conditional} with a few changes. The model is convolutional, comprised primarily of ResNet~\cite{He2016} blocks. In contrast to the standard U-Net~\cite{ronneberger2015u}, additional spatial self-attention blocks, which are essential components in transformers, are interspersed at lower resolutions. Dhariwal and Nichol~\cite{dhariwal2021adm} ablated several architecture choices for the U-Net, such as the use of adaptive normalization layers~\cite{perez2018film} to inject conditional information and channel counts for convolutional layers. However, the high-level design of the U-Net from Ho \etal has largely remained intact.

With this work, we aim to demystify the significance of architectural choices in diffusion models and offer empirical baselines for future generative modeling research. We show that the U-Net inductive bias is \emph{not} crucial to the performance of diffusion models, and they can be readily replaced with standard designs such as transformers. As a result, diffusion models are well-poised to benefit from the recent trend of architecture unification---e.g., by inheriting best practices and training recipes from other domains, as well as retaining favorable properties like scalability, robustness and efficiency. A standardized architecture would also open up new possibilities for cross-domain research.

In this paper, we focus on a new class of diffusion models based on transformers. We call them \textit{Diffusion Transformers}, or DiTs for short. DiTs adhere to the best practices of Vision Transformers (ViTs)~\cite{Dosovitskiy2020}, which have been shown to scale more effectively for visual recognition than traditional convolutional networks (e.g., ResNet~\cite{He2016}).

More specifically, we study the scaling behavior of transformers with respect to \emph{network complexity \vs sample quality}. We show that by constructing and benchmarking the DiT design space under the \emph{Latent Diffusion Models} (LDMs)~\cite{rombach2021highresolution} framework, where diffusion models are trained within a VAE's latent space, we can successfully replace the U-Net backbone with a transformer. We further show that DiTs are scalable architectures for diffusion models: there is a strong correlation between the network complexity (measured by Gflops) \vs sample quality (measured by FID). By simply scaling-up DiT and training an LDM with a high-capacity backbone (118.6 Gflops), we are able to achieve a state-of-the-art result of 2.27 FID on the class-conditional $256\times 256$ ImageNet generation benchmark. 
\vspace{-2mm}

\begin{figure*}\centering
\includegraphics[width=\linewidth]{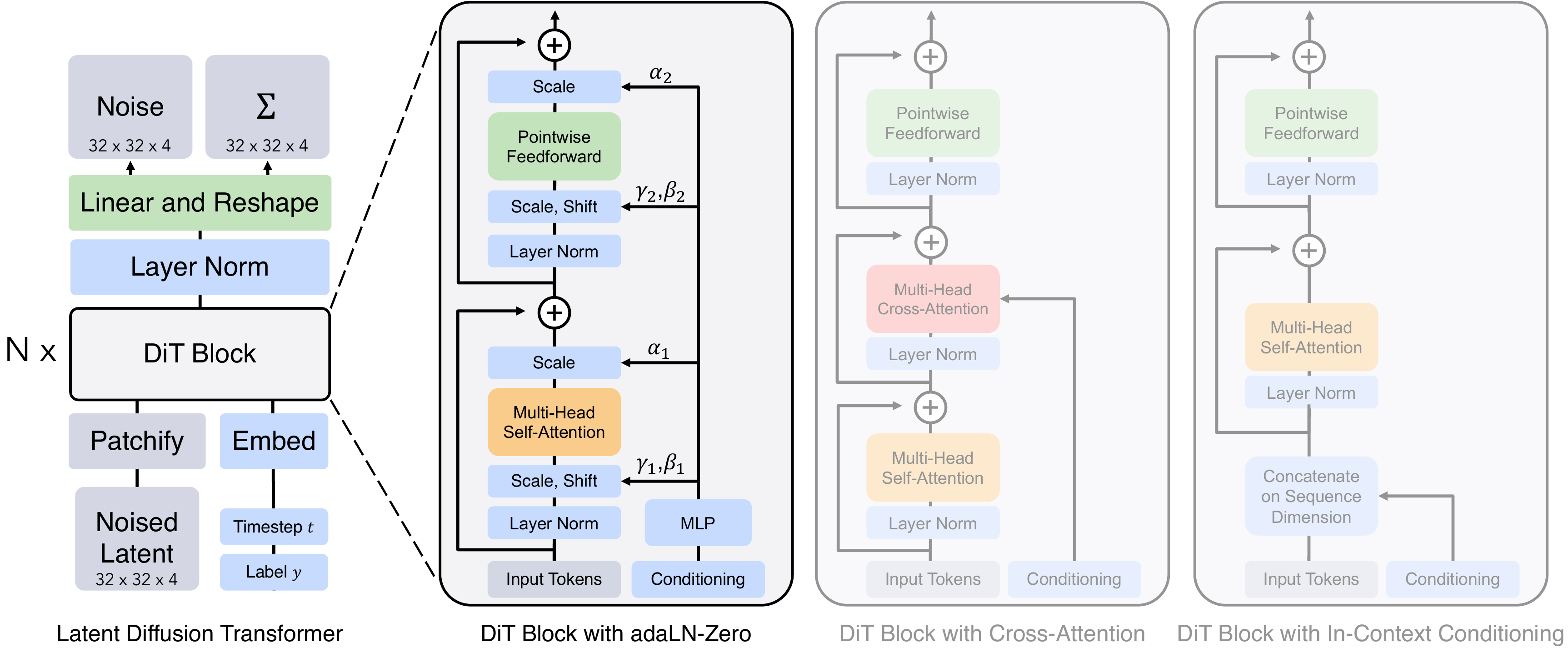}
\caption{\textbf{The Diffusion Transformer (DiT) architecture.} \emph{Left:} We train conditional latent DiT models. The input latent is decomposed into patches and processed by several DiT blocks. \emph{Right:} Details of our DiT blocks. We experiment with variants of standard transformer blocks that incorporate conditioning via adaptive layer norm, cross-attention and extra input tokens. Adaptive layer norm works best.}\vspace{-2mm}
\label{fig:architecture}\vspace{-2mm}
\end{figure*}

\section{Related Work}

\paragraph{Transformers.} Transformers~\cite{Vaswani2017} have replaced domain-specific architectures across language, vision~\cite{Dosovitskiy2020}, reinforcement learning~\cite{chen2021decision,janner2021trajectory} and meta-learning~\cite{Peebles2022}. They have shown remarkable scaling properties under increasing model size, training compute and data in the language domain~\cite{kaplan2020scaling}, as generic autoregressive models~\cite{henighan2020scaling} and as ViTs~\cite{zhai2022scaling}. Beyond language, transformers have been trained to autoregressively predict pixels~\cite{parmar2018image,child2019generating,Chen2020generative}. They have also been trained on discrete codebooks~\cite{van2017neural} as both autoregressive models~\cite{esser2020taming,ramesh2021zero} and masked generative models~\cite{chang2022maskgit,gu2022vector}; the former has shown excellent scaling behavior up to 20B parameters~\cite{yu2022scaling}. Finally, transformers have been explored in DDPMs to synthesize non-spatial data; e.g., to generate CLIP image embeddings in DALL$\cdot$E 2~\cite{ramesh2022hierarchical,radford2021}. In this paper, we study the scaling properties of transformers when used as the backbone of diffusion models of images.

\paragraph{Denoising diffusion probabilistic models (DDPMs).}
Diffusion~\cite{sohl2015thermodynamics,ho2020ddpm} and score-based generative models~\cite{hyvarinen2005estimation,song2019generative} have been particularly successful as generative models of images~\cite{nichol2021glide,ramesh2022hierarchical,imagen2022,rombach2021highresolution}, in many cases outperforming generative adversarial networks (GANs)~\cite{goodfellow2014generative} which had previously been state-of-the-art. Improvements in DDPMs over the past two years have largely been driven by improved sampling techniques~\cite{ho2020ddpm,song2020denoising,Karras2022edm}, most notably classifier-free guidance~\cite{ho2021classifier}, reformulating diffusion models to predict noise instead of pixels~\cite{ho2020ddpm} and using cascaded DDPM pipelines where low-resolution base diffusion models are trained in parallel with upsamplers~\cite{ho2021cascaded,dhariwal2021adm}. For all the diffusion models listed above, convolutional U-Nets~\cite{ronneberger2015u} are the de-facto choice of backbone architecture. Concurrent work~\cite{jabri2022scalable} introduced a novel, efficient architecture based on attention for DDPMs; we explore pure transformers.

\vspace{-4mm}
\paragraph{Architecture complexity.}
When evaluating architecture complexity in the image generation literature, it is fairly common practice to use parameter counts. In general, parameter counts can be poor proxies for the complexity of image models since they do not account for, e.g., image resolution which significantly impacts performance~\cite{radosavovic2019network,radosavovic2020designing}. Instead, much of the model complexity analysis in this paper is through the lens of theoretical Gflops. This brings us in-line with the architecture design literature where Gflops are widely-used to gauge complexity. In practice, the golden complexity metric is still up for debate as it frequently depends on particular application scenarios. Nichol and Dhariwal's seminal work improving diffusion models~\cite{nichol2021improved,dhariwal2021adm} is most related to us---there, they analyzed the scalability and Gflop properties of the U-Net architecture class. In this paper, we focus on the transformer class.

\section{Diffusion Transformers}

\subsection{Preliminaries}

\paragraph{Diffusion formulation.} Before introducing our architecture, we briefly review some basic concepts needed to understand diffusion models (DDPMs)~\cite{sohl2015thermodynamics,ho2020ddpm}. Gaussian diffusion models assume a forward noising process which gradually applies noise to real data $x_0$: $q(x_t|x_0) = \mathcal{N}(x_t; \sqrt{\bar{\alpha}_t}x_0, (1 - \bar{\alpha}_t)\mathbf{I})$, where constants $\bar{\alpha}_t$ are hyperparameters. By applying the reparameterization trick, we can sample $x_t = \sqrt{\bar{\alpha}_t}x_0 + \sqrt{1 - \bar{\alpha}_t} \epsilon_t$, where $\epsilon_t\sim\mathcal{N}(0,\mathbf{I})$.

Diffusion models are trained to learn the reverse process that inverts forward process corruptions: $p_\theta(x_{t-1}|x_t) = \mathcal{N}(\mu_\theta(x_t), \Sigma_\theta(x_t))$, where neural networks are used to predict the statistics of $p_\theta$. The reverse process model is trained with the variational lower bound~\cite{kingma2013auto} of the log-likelihood of $x_0$, which reduces to $\mathcal{L}(\theta) = -p(x_0|x_1) + \sum_t \mathcal{D}_{KL}(q^*(x_{t-1}|x_{t},x_0)||p_\theta(x_{t-1}|x_t))$, excluding an additional term irrelevant for training. Since both $q^*$ and $p_\theta$ are Gaussian, $\mathcal{D}_{KL}$ can be evaluated with the mean and covariance of the two distributions. By reparameterizing $\mu_\theta$ as a noise prediction network $\epsilon_\theta$, the model can be trained using simple mean-squared error between the predicted noise $\epsilon_\theta(x_t)$ and the ground truth sampled Gaussian noise $\epsilon_t$: $\mathcal{L}_{simple}(\theta) = ||\epsilon_\theta(x_t) - \epsilon_t||_2^2$. But, in order to train diffusion models with a learned reverse process covariance $\Sigma_\theta$, the full $\mathcal{D}_{KL}$ term needs to be optimized. We follow Nichol and Dhariwal's approach~\cite{nichol2021improved}: train $\epsilon_\theta$ with $\mathcal{L}_{simple}$, and train $\Sigma_\theta$ with the full $\mathcal{L}$. Once $p_\theta$ is trained, new images can be sampled by initializing $x_{t_\text{max}} \sim \mathcal{N}(0,\mathbf{I})$ and sampling $x_{t-1} \sim p_\theta(x_{t-1}|x_t)$ via the reparameterization trick.

\paragraph{Classifier-free guidance.} Conditional diffusion models take extra information as input, such as a class label $c$. In this case, the reverse process becomes $p_\theta(x_{t-1}|x_t,c)$, where $\epsilon_\theta$ and $\Sigma_\theta$ are conditioned on $c$. In this setting, \textit{classifier-free guidance} can be used to encourage the sampling procedure to find $x$ such that $\log p(c|x)$ is high~\cite{ho2021classifier}. By Bayes Rule, $\log p(c|x) \propto \log p(x|c) - \log p(x)$, and hence $\nabla_x \log p(c|x) \propto \nabla_x \log p(x|c) - \nabla_x \log p(x)$. By interpreting the output of diffusion models as the score function, the DDPM sampling procedure can be guided to sample $x$ with high $p(x|c)$ by:
$\hat{\epsilon}_\theta(x_t, c) = \epsilon_\theta(x_t,\emptyset) + s \cdot \nabla_x \log p(x|c) \propto \epsilon_\theta(x_t, \emptyset) + s \cdot (\epsilon_\theta(x_t, c) - \epsilon_\theta(x_t, \emptyset))$, where $s > 1$ indicates the scale of the guidance (note that $s=1$ recovers standard sampling). Evaluating the diffusion model with $c=\emptyset$ is done by randomly dropping out $c$ during training and replacing it with a learned ``null" embedding $\emptyset$. Classifier-free guidance is widely-known to yield significantly improved samples over generic sampling techniques~\cite{ho2021classifier,nichol2021glide,ramesh2022hierarchical}, and the trend holds for our DiT models.
\paragraph{Latent diffusion models.} 
Training diffusion models directly in high-resolution pixel space can be computationally prohibitive. \textit{Latent diffusion models} (LDMs)~\cite{rombach2021highresolution} tackle this issue with a two-stage approach: (1) learn an autoencoder that compresses images into smaller spatial representations with a learned encoder $E$; (2) train a diffusion model of representations $z = E(x)$ instead of a diffusion model of images $x$ ($E$ is frozen). New images can then be generated by sampling a representation $z$ from the diffusion model and subsequently decoding it to an image with the learned decoder $x = D(z)$. 

As shown in Figure~\ref{fig:bubbles}, LDMs achieve good performance while using a fraction of the Gflops of pixel space diffusion models like ADM. Since we are concerned with compute efficiency, this makes them an appealing starting point for architecture exploration. In this paper, we apply DiTs to latent space, although they could be applied to pixel space without modification as well. This makes our image generation pipeline a hybrid-based approach; we use off-the-shelf convolutional VAEs and transformer-based DDPMs.

\subsection{Diffusion Transformer Design Space}

We introduce Diffusion Transformers (DiTs), a new architecture for diffusion models. We aim to be as faithful to the standard transformer architecture as possible to retain its scaling properties. Since our focus is training DDPMs of images (specifically, spatial representations of images), DiT is based on the Vision Transformer (ViT) architecture which operates on sequences of patches~\cite{Dosovitskiy2020}. DiT retains many of the best practices of ViTs. Figure~\ref{fig:architecture} shows an overview of the complete DiT architecture. In this section, we describe the forward pass of DiT, as well as the components of the design space of the DiT class.

\begin{figure}\centering
\includegraphics[width=0.9\linewidth]{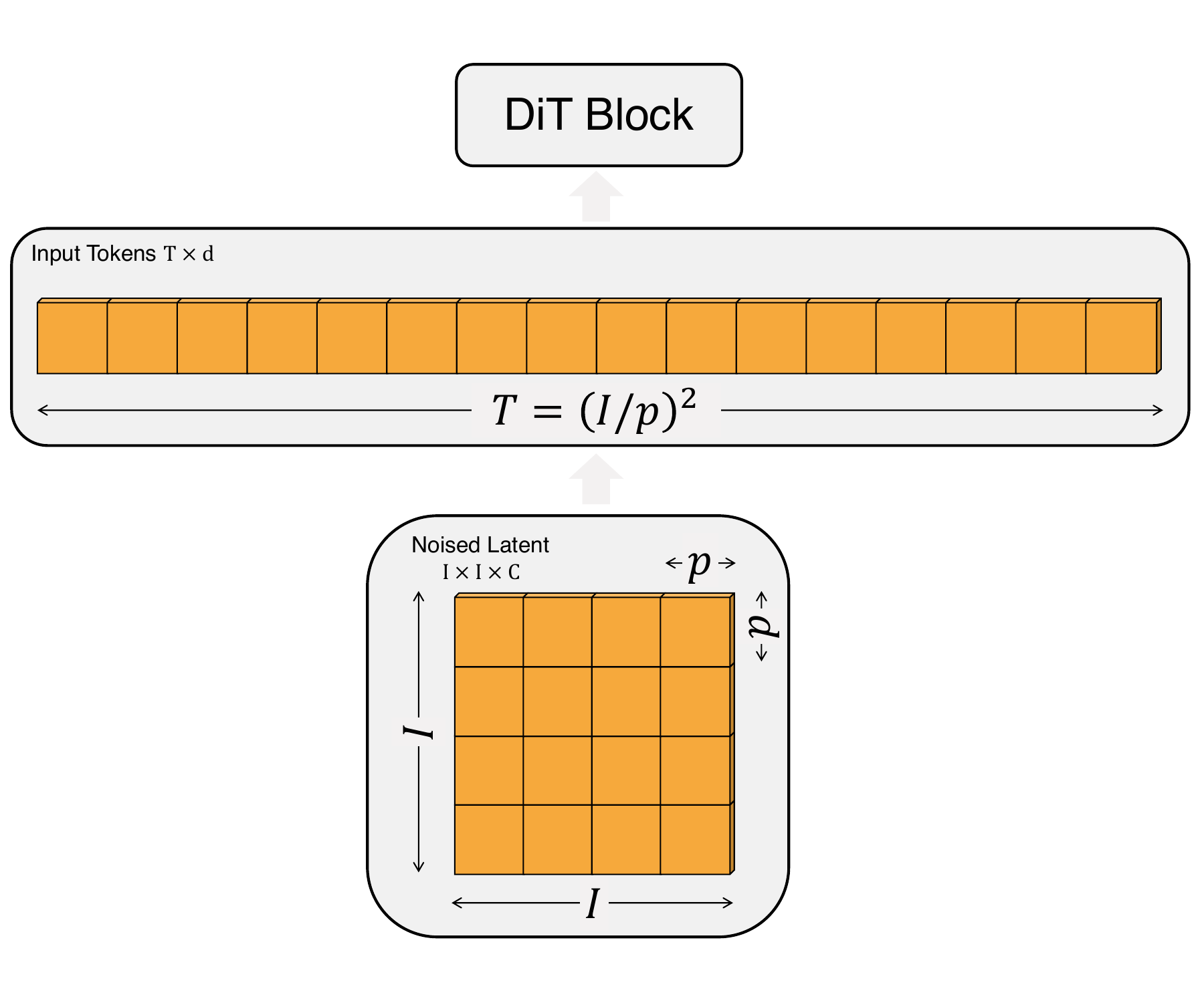}
\caption{\textbf{Input specifications for DiT.} Given patch size $p\times p$, a spatial representation (the noised latent from the VAE) of shape $I\times I \times C$ is ``patchified" into a sequence of length $T = (I/p)^2$ with hidden dimension $d$. A smaller patch size $p$ results in a longer sequence length and thus more Gflops.}\vspace{-2mm}
\label{fig:patchify}
\end{figure}
\vspace{-2mm}
\paragraph{Patchify.} The input to DiT is a spatial representation $z$ (for $256\times256\times3$ images, $z$ has shape $32\times32\times4$). The first layer of DiT is ``patchify," which converts the spatial input into a sequence of $T$ tokens, each of dimension $d$, by linearly embedding each patch in the input. Following patchify, we apply standard ViT frequency-based positional embeddings (the sine-cosine version) to all input tokens. The number of tokens $T$ created by patchify is determined by the patch size hyperparameter $p$. As shown in Figure~\ref{fig:patchify}, halving $p$ will quadruple $T$, and thus \textit{at least} quadruple total transformer Gflops. Although it has a significant impact on Gflops, note that changing $p$ has no meaningful impact on downstream parameter counts. 

\emph{We add $p=2,4,8$ to the DiT design space.} 
\vspace{-2mm}
\paragraph{DiT block design.} Following patchify, the input tokens are processed by a sequence of transformer blocks. In addition to noised image inputs, diffusion models sometimes process additional conditional information such as noise timesteps $t$, class labels $c$, natural language, etc. We explore four variants of transformer blocks that process conditional inputs differently. The designs introduce small, but important, modifications to the standard ViT block design. The designs of all blocks are shown in Figure~\ref{fig:architecture}.

\begin{figure}\centering
\includegraphics[width=0.9\linewidth]{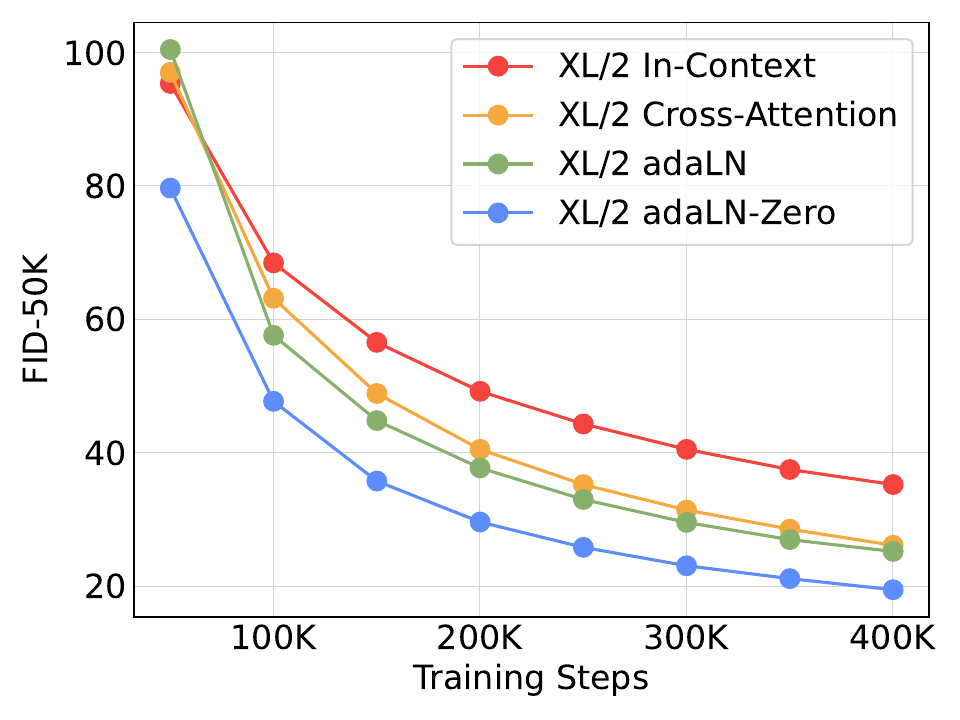}
\caption{\textbf{Comparing different conditioning strategies.} adaLN-Zero outperforms cross-attention and in-context conditioning at all stages of training.}\vspace{-4mm}
\label{fig:conditioning}
\end{figure}

\begin{itemize}
\item[--] \textit{In-context conditioning.} We simply append the vector embeddings of $t$ and $c$ as two additional tokens in the input sequence, treating them no differently from the image tokens. This is similar to \texttt{cls} tokens in ViTs, and it allows us to use standard ViT blocks without modification. After the final block, we remove the conditioning tokens from the sequence. This approach introduces negligible new Gflops to the model.

\item[--] \textit{Cross-attention block.} We concatenate the embeddings of $t$ and $c$ into a length-two sequence, separate from the image token sequence. The transformer block is modified to include an additional multi-head cross-attention layer following the multi-head self-attention block, similar to the original design from Vaswani \etal \cite{Vaswani2017}, and also similar to the one used by LDM for conditioning on class labels. Cross-attention adds the most Gflops to the model, roughly a 15\% overhead.

\item[--] \textit{Adaptive layer norm (adaLN) block.} Following the widespread usage of adaptive normalization layers~\cite{perez2018film} in GANs~\cite{brock2018large,karras2019style} and diffusion models with U-Net backbones~\cite{dhariwal2021adm}, we explore replacing standard layer norm layers in transformer blocks with adaptive layer norm (adaLN). Rather than directly learn dimension-wise scale and shift parameters $\gamma$ and $\beta$, we regress them from the sum of the embedding vectors of $t$ and $c$. Of the three block designs we explore, adaLN adds the least Gflops and is thus the most compute-efficient. It is also the only conditioning mechanism that is restricted to apply the \emph{same function} to all tokens.

\item[--] \textit{adaLN-Zero block.} Prior work on ResNets has found that initializing each residual block as the identity function is beneficial. For example, Goyal \etal found that zero-initializing the final batch norm scale factor $\gamma$ in each block accelerates large-scale training in the supervised learning setting~\cite{Goyal2017b}. Diffusion U-Net models use a similar initialization strategy, zero-initializing the final convolutional layer in each block prior to any residual connections. We explore a modification of the adaLN DiT block which does the same. In addition to regressing $\gamma$ and $\beta$, we also regress dimension-wise scaling parameters $\alpha$ that are applied immediately prior to any residual connections within the DiT block. We initialize the MLP to output the zero-vector for all $\alpha$; this initializes the full DiT block as the identity function. As with the vanilla adaLN block, adaLN-Zero adds negligible Gflops to the model.

\end{itemize}

\emph{We include the in-context, cross-attention, adaptive layer norm and adaLN-Zero blocks in the DiT design space.} 

\begin{table}
\centering
\small
\scalebox{0.9}{
\begin{tabular}{l c c c c c}
\toprule
Model            & Layers $N$ & Hidden size $d$ &  Heads  & Gflops \tiny{($I$=32, $p$=4)} \\
\midrule 
DiT-S   &   12   &     384   &   6    &   1.4  \\
DiT-B   &   12   &      768    &   12   &   5.6  \\
DiT-L  &    24   &      1024    &   16   &  19.7  \\
DiT-XL &    28  &       1152     &   16  &  29.1  \\
\bottomrule
\end{tabular}}
\caption{\textbf{Details of DiT models.} We follow ViT~\cite{Dosovitskiy2020} model configurations for the Small (S), Base (B) and Large (L) variants; we also introduce an XLarge (XL) config as our largest model.}
\label{tbl:models}
\end{table}

\paragraph{Model size.}
We apply a sequence of $N$ DiT blocks, each operating at the hidden dimension size $d$. Following ViT, we use standard transformer configs that jointly scale $N$, $d$ and attention heads~\cite{Dosovitskiy2020,zhai2022scaling}. Specifically, we use four configs: DiT-S, DiT-B, DiT-L and DiT-XL. They cover a wide range of model sizes and flop allocations, from 0.3 to 118.6 Gflops, allowing us to gauge scaling performance. Table~\ref{tbl:models} gives details of the configs. 

\textit{We add B, S, L and XL configs to the DiT design space.}

\begin{figure*}\centering
\includegraphics[width=\linewidth]{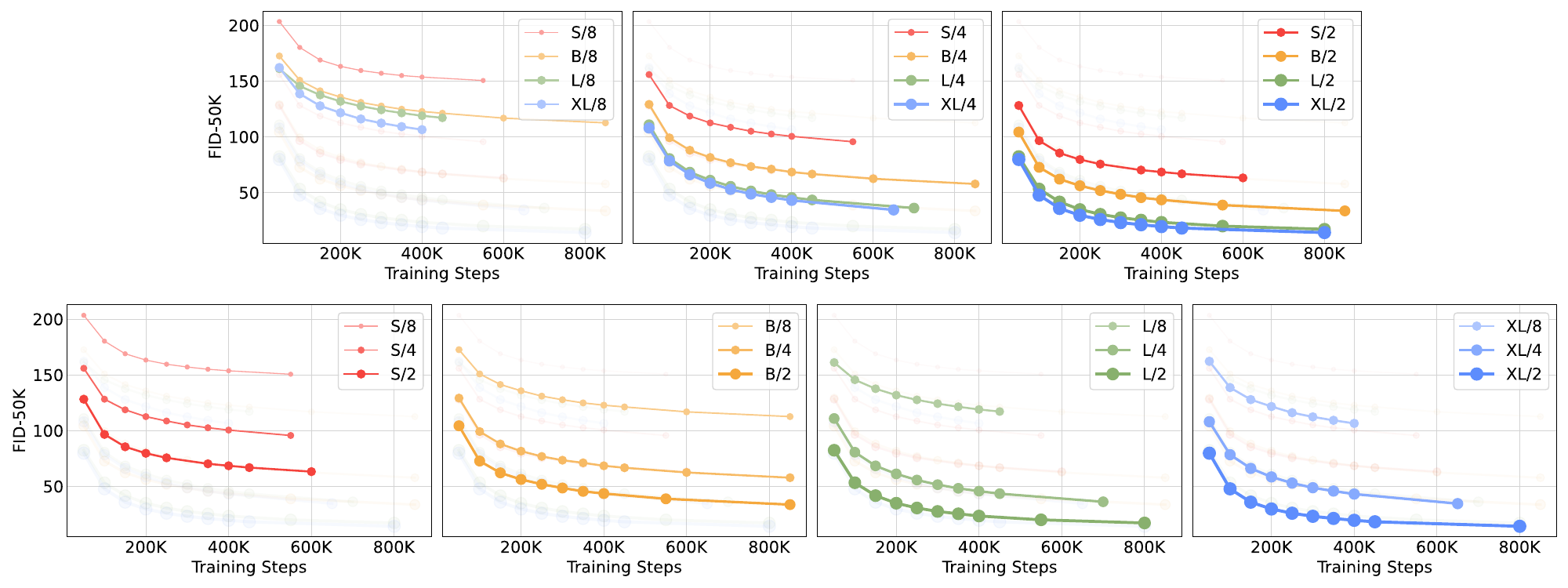}
\caption{\textbf{Scaling the DiT model improves FID at all stages of training.} We show FID-50K over training iterations for 12 of our DiT models. \emph{Top row:} We compare FID holding patch size constant. \emph{Bottom row:} We compare FID holding model size constant. Scaling the transformer backbone yields better generative models across all model sizes and patch sizes.}\vspace{-2mm}
\label{fig:scaling}
\end{figure*}

\vspace{-2mm}
\paragraph{Transformer decoder.}
After the final DiT block, we need to decode our sequence of image tokens into an output noise prediction and an output diagonal covariance prediction. Both of these outputs have shape equal to the original spatial input. We use a standard linear decoder to do this; we apply the final layer norm (adaptive if using adaLN) and linearly decode each token into a $p \times p \times 2C$ tensor, where $C$ is the number of channels in the spatial input to DiT. Finally, we rearrange the decoded tokens into their original spatial layout to get the predicted noise and covariance.

\textit{The complete DiT design space we explore is patch size, transformer block architecture and model size.}

\section{Experimental Setup}

We explore the DiT design space and study the scaling properties of our model class. Our models are named according to their configs and latent patch sizes $p$; for example, DiT-XL/2 refers to the XLarge config and $p=2$.
\paragraph{Training.} We train class-conditional latent DiT models at $256\times256$ and $512\times512$ image resolution on the ImageNet dataset~\cite{krizhevsky2012imagenet}, a highly-competitive generative modeling benchmark. We initialize the final linear layer with zeros and otherwise use standard weight initialization techniques from ViT. We train all models with AdamW~\cite{loshchilov2017decoupled,kingma2014adam}. We use a constant learning rate of $1 \times 10^{-4}$, no weight decay and a batch size of 256. The only data augmentation we use is horizontal flips. Unlike much prior work with ViTs~\cite{steiner2021train,Xiao2021}, we did not find learning rate warmup nor regularization necessary to train DiTs to high performance. Even without these techniques, training was highly stable across all model configs and we did not observe any loss spikes commonly seen when training transformers. Following common practice in the generative modeling literature, we maintain an exponential moving average (EMA) of DiT weights over training with a decay of 0.9999. All results reported use the EMA model. We use identical training hyperparameters across all DiT model sizes and patch sizes. Our training hyperparameters are almost entirely retained from ADM. \emph{We did \textit{not} tune learning rates, decay/warm-up schedules, Adam $\beta_1$/$\beta_2$ or weight decays.}
\vspace{-2mm}
\paragraph{Diffusion.} We use an off-the-shelf pre-trained variational autoencoder (VAE) model~\cite{kingma2013auto} from Stable Diffusion~\cite{rombach2021highresolution}. The VAE encoder has a downsample factor of 8---given an RGB image $x$ with shape $256 \times 256 \times 3$, $z = E(x)$ has shape $32 \times 32 \times 4$. Across all experiments in this section, our diffusion models operate in this $\mathcal{Z}$-space. After sampling a new latent from our diffusion model, we decode it to pixels using the VAE decoder $x = D(z)$. We retain diffusion hyperparameters from ADM~\cite{dhariwal2021adm}; specifically, we use a $t_\text{max}=1000$ linear variance schedule ranging from $1 \times 10^{-4}$ to $2 \times 10^{-2}$, ADM's parameterization of the covariance $\Sigma_\theta$ and their method for embedding input timesteps and labels.
\vspace{-6mm}
\paragraph{Evaluation metrics.} We measure scaling performance with Fréchet Inception Distance (FID)~\cite{heusel2017gans}, the standard metric for evaluating generative models of images. 

\noindent We follow convention when comparing against prior works and report FID-50K using 250 DDPM sampling steps. FID is known to be sensitive to small implementation details~\cite{parmar2021cleanfid}; to ensure accurate comparisons, all values reported in this paper are obtained by exporting samples and using ADM's TensorFlow evaluation suite~\cite{dhariwal2021adm}. FID numbers reported in this section do \textit{not} use classifier-free guidance except where otherwise stated. We additionally report Inception Score~\cite{salimans2016gans}, sFID~\cite{nash2021generating} and Precision/Recall~\cite{kynkaanniemi2019improved} as secondary metrics.
\paragraph{Compute.} We implement all models in JAX~\cite{jax2018github} and train them using TPU-v3 pods. DiT-XL/2, our most compute-intensive model, trains at roughly 5.7 iterations/second on a TPU v3-256 pod with a global batch size of 256. 

\section{Experiments}\label{sec:experiments}

\paragraph{DiT block design.} We train four of our highest Gflop DiT-XL/2 models, each using a different block design---in-context (119.4 Gflops), cross-attention (137.6 Gflops), adaptive layer norm (adaLN, 118.6 Gflops) or adaLN-zero (118.6 Gflops). We measure FID over the course of training. Figure~\ref{fig:conditioning} shows the results. The adaLN-Zero block yields lower FID than both cross-attention and in-context conditioning while being the most compute-efficient. At 400K training iterations, the FID achieved with the adaLN-Zero model is nearly half that of the in-context model, demonstrating that the conditioning mechanism critically affects model quality. Initialization is also important---adaLN-Zero, which initializes each DiT block as the identity function, significantly outperforms vanilla adaLN. \textit{For the rest of the paper, all models will use adaLN-Zero DiT blocks.}

\begin{figure*}\centering\vspace{-4mm}%
\includegraphics[width=\linewidth]{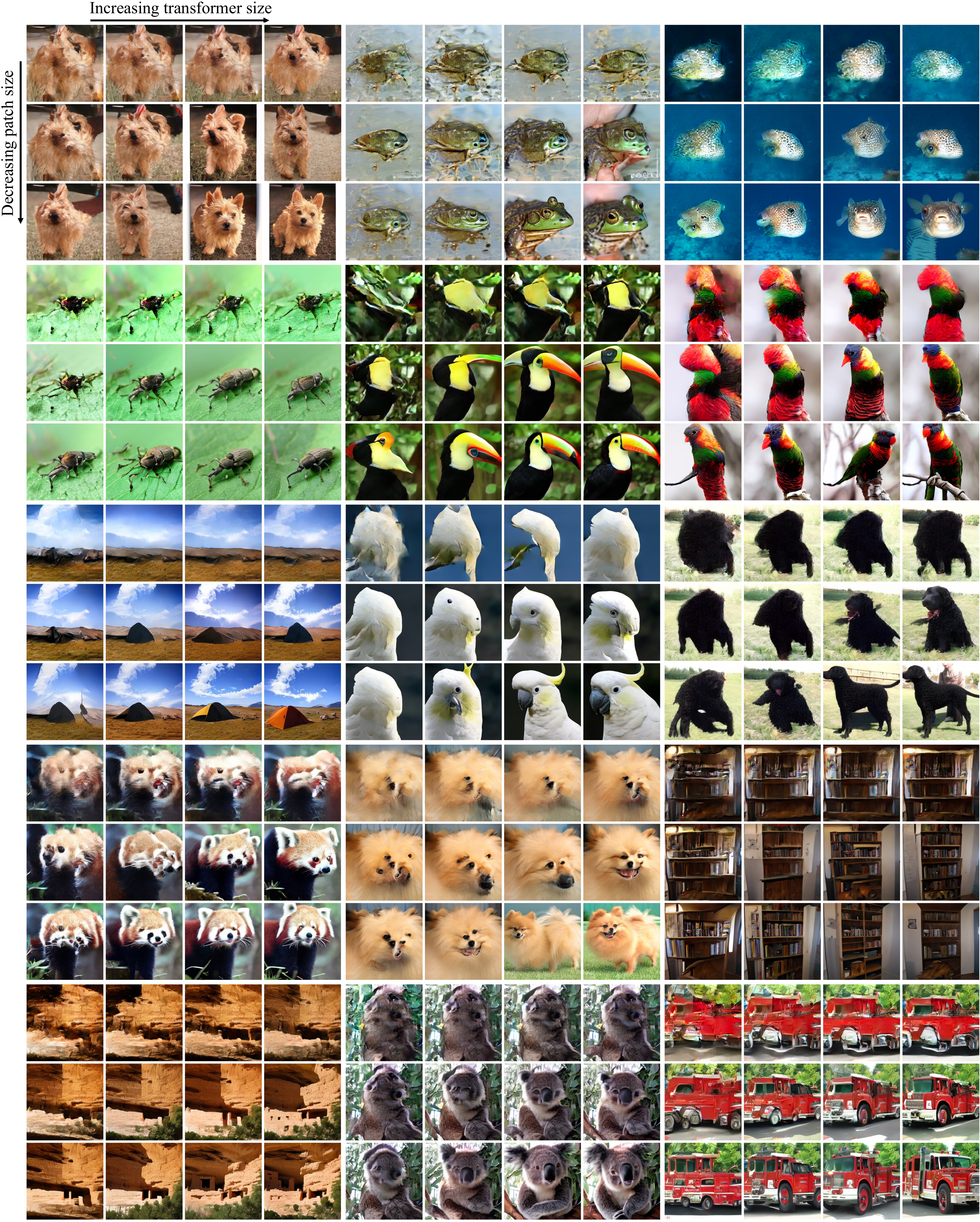}
\vspace{-2mm}
\caption{\textbf{Increasing transformer forward pass Gflops increases sample quality.} \textit{Best viewed zoomed-in.} We sample from all 12 of our DiT models after 400K training steps using the same input latent noise and class label. Increasing the Gflops in the model---either by increasing transformer depth/width or increasing the number of input tokens---yields significant improvements in visual fidelity.}
\label{fig:visual_scaling}
\end{figure*}

\vspace{-2mm}
\paragraph{Scaling model size and patch size.}
We train 12 DiT models, sweeping over model configs (S, B, L, XL) and patch sizes (8, 4, 2). Note that DiT-L and DiT-XL are significantly closer to each other in terms of relative Gflops than other configs. Figure~\ref{fig:bubbles} (left) gives an overview of the Gflops of each model and their FID at 400K training iterations. In all cases, we find that increasing model size and decreasing patch size yields considerably improved diffusion models.

\begin{figure}\centering\vspace{-0.85mm}
\includegraphics[width=\linewidth]{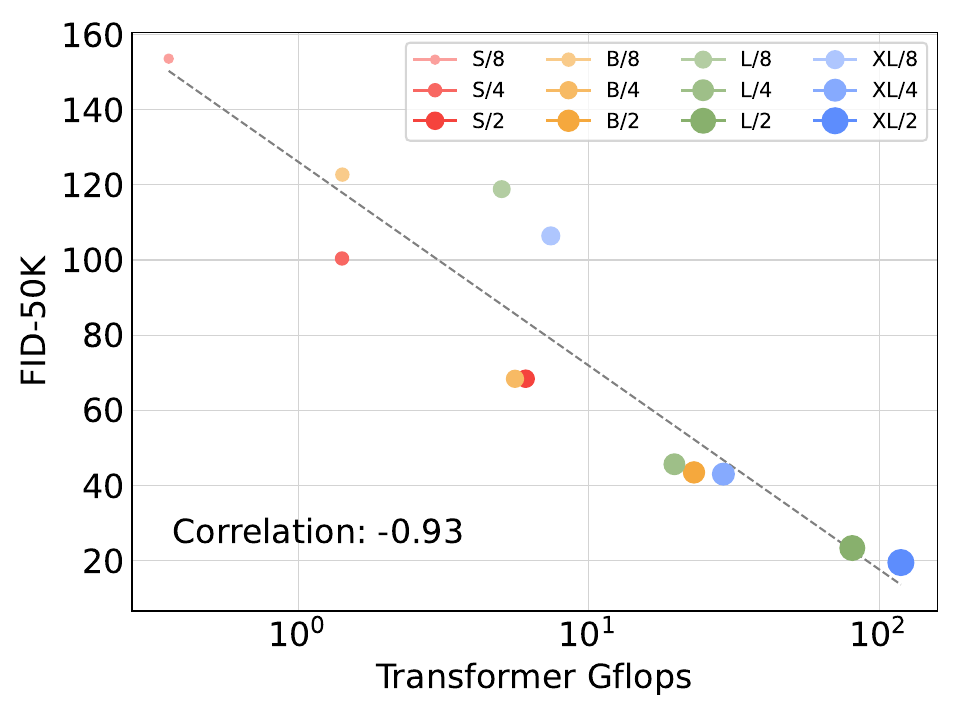}
\vspace{-5.4mm}
\caption{\textbf{Transformer Gflops are strongly correlated with FID.} We plot the Gflops of each of our DiT models and each model's FID-50K after 400K training steps.}\vspace{-4mm}
\label{fig:gflop_line}
\end{figure}

Figure~\ref{fig:scaling} (top) demonstrates how FID changes as model size is increased and patch size is held constant. Across all four configs, significant improvements in FID are obtained over all stages of training by making the transformer deeper and wider. Similarly, Figure~\ref{fig:scaling} (bottom) shows FID as patch size is decreased and model size is held constant. We again observe considerable FID improvements throughout training by simply scaling the number of tokens processed by DiT, holding parameters approximately fixed.

\vspace{-3mm}
\paragraph{DiT Gflops are critical to improving performance.}

The results of Figure~\ref{fig:scaling} suggest that parameter counts do not uniquely determine the quality of a DiT model. As model size is held constant and patch size is decreased, the transformer's total parameters are effectively unchanged (actually, total parameters slightly \textit{decrease}), and only Gflops are increased. These results indicate that scaling model \textit{Gflops} is actually the key to improved performance. To investigate this further, we plot the FID-50K at 400K training steps against model Gflops in Figure~\ref{fig:gflop_line}. The results demonstrate that different DiT configs obtain similar FID values when their total Gflops are similar (e.g., DiT-S/2 and DiT-B/4). We find a strong negative correlation between model Gflops and FID-50K, suggesting that additional model compute is the critical ingredient for improved DiT models. In Figure~\ref{fig:training_complexity_ISPR} (appendix), we find that this trend holds for other metrics such as Inception Score.

\begin{figure}\centering
\includegraphics[width=\linewidth]{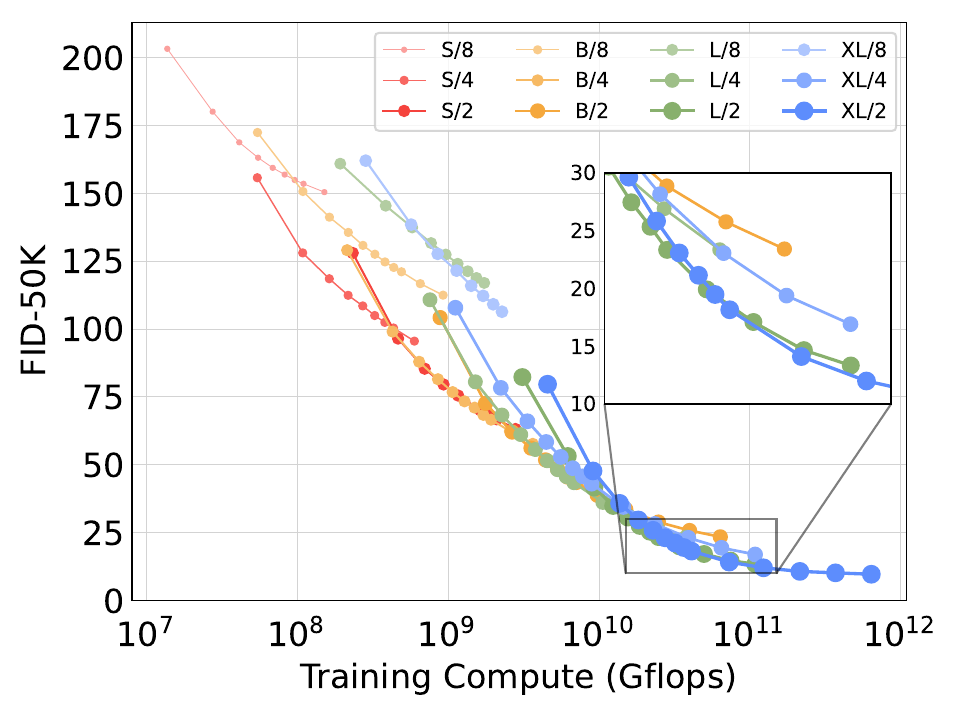}
\vspace{-6.25mm}
\caption{\textbf{Larger DiT models use large compute more efficiently.} We plot FID as a function of total training compute.}
\label{fig:training_complexity_fid}
\end{figure}

\textbf{Larger DiT models are more compute-efficient.} In Figure~\ref{fig:training_complexity_fid}, we plot FID as a function of total training compute for all DiT models. We estimate training compute as model Gflops $\cdot$ batch size $\cdot$ training steps $\cdot$ 3, where the factor of 3 roughly approximates the backwards pass as being twice as compute-heavy as the forward pass. We find that small DiT models, even when trained longer, eventually become compute-inefficient relative to larger DiT models trained for fewer steps. Similarly, we find that models that are identical except for patch size have different performance profiles even when controlling for training Gflops. For example, XL/4 is outperformed by XL/2 after roughly $10^{10}$ Gflops.

\vspace{-2mm}
\paragraph{Visualizing scaling.} We visualize the effect of scaling on sample quality in Figure~\ref{fig:visual_scaling}. At 400K training steps, we sample an image from each of our 12 DiT models using \textit{identical} starting noise $x_{t_\text{max}}$, sampling noise and class labels. This lets us visually interpret how scaling affects DiT sample quality. Indeed, scaling both model size and the number of tokens yields notable improvements in visual quality. 
\vspace{2mm}
\subsection{State-of-the-Art Diffusion Models}

\paragraph{256$\times$256 ImageNet.} Following our scaling analysis, we continue training our highest Gflop model, DiT-XL/2, for 7M steps. We show samples from the model in Figures~\ref{fig:teaser}, and we compare against state-of-the-art class-conditional generative models. We report results in Table~\ref{tbl:sota}. When using classifier-free guidance, DiT-XL/2 outperforms all prior diffusion models, decreasing the previous best FID-50K of 3.60 achieved by LDM to 2.27. Figure~\ref{fig:bubbles} (right) shows that DiT-XL/2 (118.6 Gflops) is compute-efficient relative to latent space U-Net models like LDM-4 (103.6 Gflops) and substantially more efficient than pixel space U-Net models such as ADM (1120 Gflops) or ADM-U (742 Gflops). Our method achieves the lowest FID of all prior generative models, including the previous state-of-the-art StyleGAN-XL~\cite{Sauer2021ARXIV}. Finally, we also observe that DiT-XL/2 achieves higher recall values at all tested classifier-free guidance scales compared to LDM-4 and LDM-8. When trained for only 2.35M steps (similar to ADM), XL/2 still outperforms all prior diffusion models with an FID of 2.55. %

\begin{table}[t]
    \begin{center}
    \begin{small}
    \scalebox{0.81}{
    \begin{tabular}{lccccc}
    \toprule
    \multicolumn{6}{l}{\bf{Class-Conditional ImageNet} 256$\times$256} \\
    \toprule
    Model & FID$\downarrow$   & sFID$\downarrow$  & IS$\uparrow$     & Precision$\uparrow$ & Recall$\uparrow$ \\
    \bottomrule
    BigGAN-deep~\cite{brock2018large} & 6.95 & 7.36 & 171.4 & 0.87 & 0.28 \\
    StyleGAN-XL~\cite{Sauer2021ARXIV} & 2.30 & 4.02 & 265.12 & 0.78 & 0.53 \\
    \toprule
    ADM~\cite{dhariwal2021adm} & 10.94 & 6.02 & 100.98 & 0.69 & 0.63 \\
    ADM-U & 7.49 & 5.13 & 127.49 & 0.72 & 0.63 \\
    ADM-G & 4.59 & 5.25 & 186.70 & 0.82 & 0.52 \\
    ADM-G, ADM-U & 3.94 & 6.14      & 215.84 & 0.83 & 0.53 \\
    \arrayrulecolor{gray}\cmidrule(lr){1-6}
    CDM~\cite{ho2021cascaded} & 4.88 & - & 158.71 & - & - \\
    \arrayrulecolor{gray}\cmidrule(lr){1-6}
    LDM-8~\cite{rombach2021highresolution} & 15.51 & - & 79.03 & 0.65 & 0.63 \\
    LDM-8-G & 7.76 & - & 209.52 & 0.84 & 0.35 \\
    LDM-4 & 10.56 & - & 103.49 & 0.71 & 0.62 \\
    LDM-4-G (cfg=1.25) &  3.95 & - & 178.22 & 0.81 & 0.55 \\
    LDM-4-G (cfg=1.50) & 3.60 & - & 247.67 & \textbf{0.87} & 0.48 \\
    \arrayrulecolor{gray}\cmidrule(lr){1-6}
    \textbf{DiT-XL/2}              & 9.62 & 6.85 & 121.50 & 0.67 & \textbf{0.67} \\
    \textbf{DiT-XL/2-G} (cfg=1.25) & 3.22 & 5.28 & 201.77 & 0.76 & 0.62 \\
    \textbf{DiT-XL/2-G} (cfg=1.50) & \textbf{2.27} & \textbf{4.60} & \textbf{278.24} & 0.83 & 0.57 \\
    \bottomrule
    \end{tabular}
    } %
    \end{small}
    \end{center}
    \vspace{-4mm}
    \caption{\textbf{Benchmarking class-conditional image generation on ImageNet 256$\times$256.} DiT-XL/2 achieves state-of-the-art FID.} %
    \label{tbl:sota}
    \vskip -0.2in
    \vspace{2mm}
    \begin{center}
    \begin{small}
    \scalebox{0.81}{
    \begin{tabular}{lccccc}
    \toprule
    \multicolumn{6}{l}{\bf{Class-Conditional ImageNet} 512$\times$512} \\
    \toprule
    Model & FID$\downarrow$   & sFID$\downarrow$  & IS$\uparrow$     & Precision$\uparrow$ & Recall$\uparrow$ \\
    \bottomrule
    BigGAN-deep~\cite{brock2018large} & 8.43 & 8.13 & 177.90 & 0.88 & 0.29 \\
    StyleGAN-XL~\cite{Sauer2021ARXIV} & 2.41 & 4.06 & 267.75 & 0.77 & 0.52 \\
    \toprule
    ADM~\cite{dhariwal2021adm} & 23.24 & 10.19 & 58.06 & 0.73 & 0.60  \\
    ADM-U & 9.96 & 5.62 & 121.78 & 0.75 & \textbf{0.64} \\
    ADM-G & 7.72 & 6.57 & 172.71 & \textbf{0.87} & 0.42 \\
    ADM-G, ADM-U & 3.85 & 5.86 & 221.72 & 0.84 & 0.53\\
    \arrayrulecolor{gray}\cmidrule(lr){1-6}
    \textbf{DiT-XL/2} & 12.03 & 7.12 & 105.25 & 0.75 & \textbf{0.64} \\
    \textbf{DiT-XL/2-G} (cfg=1.25) & 4.64 & 5.77 & 174.77 & 0.81 & 0.57 \\
    \textbf{DiT-XL/2-G} (cfg=1.50) & \textbf{3.04} & \textbf{5.02} & \textbf{240.82} & 0.84 & 0.54 \\
    \bottomrule
    \end{tabular}
    } %
    \end{small}
    \end{center}
    \vspace{-4mm}
    \caption{\textbf{Benchmarking class-conditional image generation on ImageNet 512$\times$512.} Note that prior work~\cite{dhariwal2021adm} measures Precision and Recall using 1000 real samples for $512\times512$ resolution; for consistency, we do the same.}
    \label{tbl:sota512}
    \vskip -0.2in
\end{table}

\begin{figure}\centering%
\includegraphics[width=\linewidth]{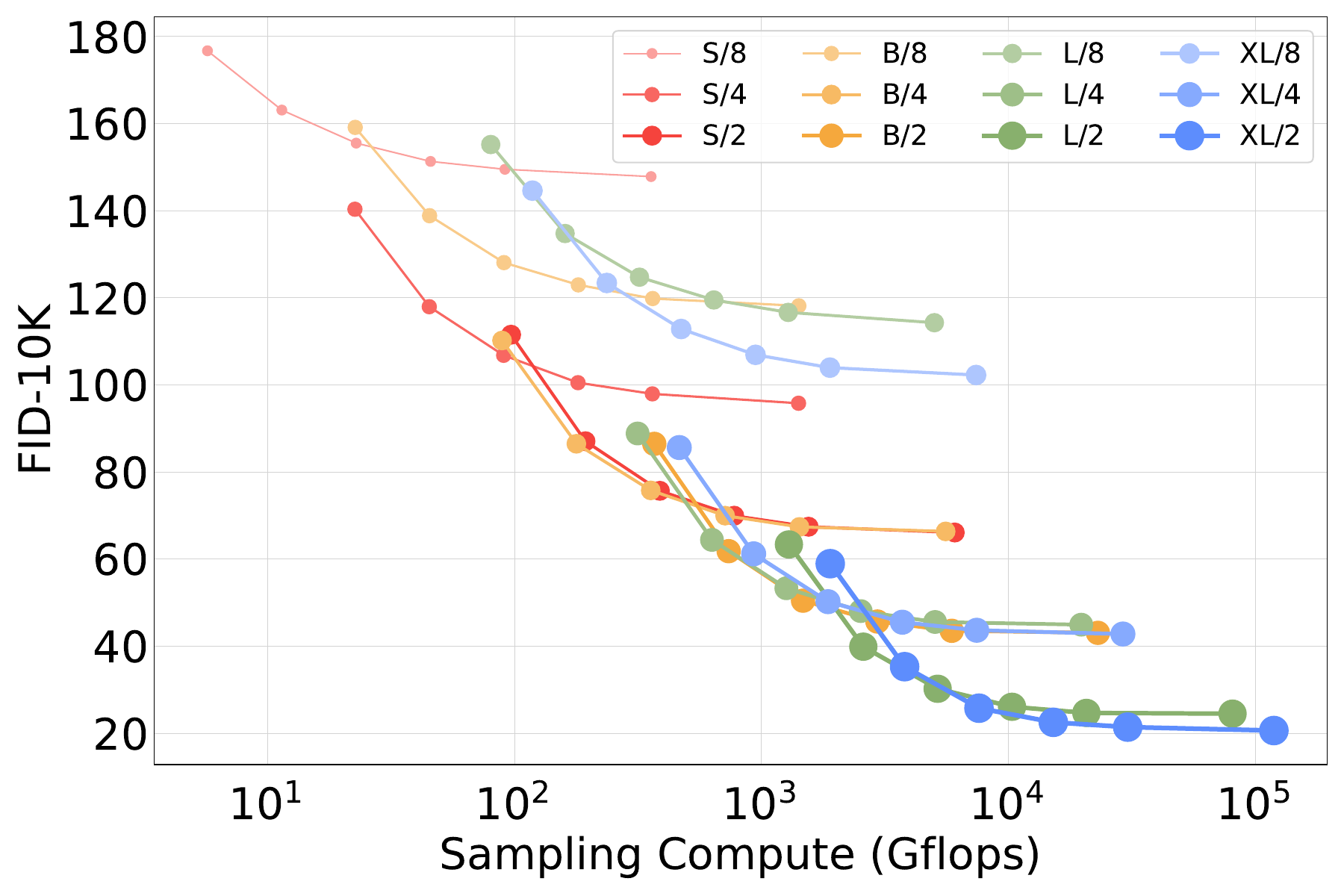}
\vspace{-4mm}
\caption{\textbf{Scaling-up \textit{sampling} compute does not compensate for a lack of \textit{model} compute.} For each of our DiT models trained for 400K iterations, we compute FID-10K using [16, 32, 64, 128, 256, 1000] sampling steps. For each number of steps, we plot the FID as well as the Gflops used to sample each image. Small models cannot close the performance gap with our large models, even if they sample with more test-time Gflops than the large models.}\vspace{-2mm}
\label{fig:sample_gflops}
\end{figure}
\vspace{-2mm}

\paragraph{512$\times$512 ImageNet.} 
We train a new DiT-XL/2 model on ImageNet at $512\times512$ resolution for 3M iterations with identical hyperparameters as the $256\times256$ model. With a patch size of 2, this XL/2 model processes a total of 1024 tokens after patchifying the $64\times64\times4$ input latent (524.6 Gflops). Table~\ref{tbl:sota512} shows comparisons against state-of-the-art methods. XL/2 again outperforms all prior diffusion models at this resolution, improving the previous best FID of 3.85 achieved by ADM to 3.04. Even with the increased number of tokens, XL/2 remains compute-efficient. For example, ADM uses 1983 Gflops and ADM-U uses 2813 Gflops; XL/2 uses 524.6 Gflops. We show samples from the high-resolution XL/2 model in Figure~\ref{fig:teaser} and the appendix.

\subsection{Scaling Model vs. Sampling Compute}
Diffusion models are unique in that they can use additional compute after training by increasing the number of sampling steps when generating an image. Given the impact of model Gflops on sample quality, in this section we study if smaller-\textit{model compute} DiTs can outperform larger ones by using more \textit{sampling compute}. We compute FID for all 12 of our DiT models after 400K training steps, using [16, 32, 64, 128, 256, 1000] sampling steps per-image. The main results are in Figure~\ref{fig:sample_gflops}. Consider DiT-L/2 using 1000 sampling steps versus DiT-XL/2 using 128 steps. In this case, L/2 uses $80.7$ Tflops to sample each image; XL/2 uses $5\times$ less compute---$15.2$ Tflops---to sample each image. Nonetheless, XL/2 has the better FID-10K (23.7 vs 25.9). In general, scaling-up sampling compute \textit{cannot} compensate for a lack of model compute.
\vspace{-1mm}

\section{Conclusion}
We introduce Diffusion Transformers (DiTs), a simple transformer-based backbone for diffusion models that outperforms prior U-Net models and inherits the excellent scaling properties of the transformer model class. Given the promising scaling results in this paper, future work should continue to scale DiTs to larger models and token counts. DiT could also be explored as a drop-in backbone for text-to-image models like DALL$\cdot$E 2 and Stable Diffusion.

\paragraph{Acknowledgements.} 
We thank Kaiming He, Ronghang Hu, Alexander Berg, Shoubhik Debnath, Tim Brooks, Ilija Radosavovic and Tete Xiao for helpful discussions. William Peebles is supported by the NSF GRFP.

{\small
\bibliographystyle{ieee_fullname}
\bibliography{egbib}

\begin{thebibliography}{10}\itemsep=-1pt

\bibitem{jax2018github}
James Bradbury, Roy Frostig, Peter Hawkins, Matthew~James Johnson, Chris Leary,
  Dougal Maclaurin, George Necula, Adam Paszke, Jake Vander{P}las, Skye
  Wanderman-{M}ilne, and Qiao Zhang.
\newblock {JAX}: composable transformations of {P}ython+{N}um{P}y programs,
  2018.

\bibitem{brock2018large}
Andrew Brock, Jeff Donahue, and Karen Simonyan.
\newblock Large scale {GAN} training for high fidelity natural image synthesis.
\newblock In {\em ICLR}, 2019.

\bibitem{Brown2020}
Tom~B Brown, Benjamin Mann, Nick Ryder, Melanie Subbiah, Jared Kaplan, Prafulla
  Dhariwal, Arvind Neelakantan, Pranav Shyam, Girish Sastry, Amanda Askell,
  et~al.
\newblock Language models are few-shot learners.
\newblock In {\em NeurIPS}, 2020.

\bibitem{chang2022maskgit}
Huiwen Chang, Han Zhang, Lu Jiang, Ce Liu, and William~T Freeman.
\newblock Maskgit: Masked generative image transformer.
\newblock In {\em CVPR}, pages 11315--11325, 2022.

\bibitem{chen2021decision}
Lili Chen, Kevin Lu, Aravind Rajeswaran, Kimin Lee, Aditya Grover, Misha
  Laskin, Pieter Abbeel, Aravind Srinivas, and Igor Mordatch.
\newblock Decision transformer: Reinforcement learning via sequence modeling.
\newblock In {\em NeurIPS}, 2021.

\bibitem{Chen2020generative}
Mark Chen, Alec Radford, Rewon Child, Jeffrey Wu, Heewoo Jun, David Luan, and
  Ilya Sutskever.
\newblock Generative pretraining from pixels.
\newblock In {\em ICML}, 2020.

\bibitem{child2019generating}
Rewon Child, Scott Gray, Alec Radford, and Ilya Sutskever.
\newblock Generating long sequences with sparse transformers.
\newblock {\em arXiv preprint arXiv:1904.10509}, 2019.

\bibitem{Devlin2019}
Jacob Devlin, Ming-Wei Chang, Kenton Lee, and Kristina Toutanova.
\newblock Bert: Pre-training of deep bidirectional transformers for language
  understanding.
\newblock In {\em NAACL-HCT}, 2019.

\bibitem{dhariwal2021adm}
Prafulla Dhariwal and Alexander Nichol.
\newblock Diffusion models beat gans on image synthesis.
\newblock In {\em NeurIPS}, 2021.

\bibitem{Dosovitskiy2020}
Alexey Dosovitskiy, Lucas Beyer, Alexander Kolesnikov, Dirk Weissenborn,
  Xiaohua Zhai, Thomas Unterthiner, Mostafa Dehghani, Matthias Minderer, Georg
  Heigold, Sylvain Gelly, et~al.
\newblock An image is worth 16x16 words: Transformers for image recognition at
  scale.
\newblock In {\em ICLR}, 2020.

\bibitem{esser2020taming}
Patrick Esser, Robin Rombach, and Björn Ommer.
\newblock Taming transformers for high-resolution image synthesis, 2020.

\bibitem{goodfellow2014generative}
Ian Goodfellow, Jean Pouget-Abadie, Mehdi Mirza, Bing Xu, David Warde-Farley,
  Sherjil Ozair, Aaron Courville, and Yoshua Bengio.
\newblock Generative adversarial nets.
\newblock In {\em NIPS}, 2014.

\bibitem{Goyal2017b}
Priya Goyal, Piotr Doll{\'a}r, Ross Girshick, Pieter Noordhuis, Lukasz
  Wesolowski, Aapo Kyrola, Andrew Tulloch, Yangqing Jia, and Kaiming He.
\newblock Accurate, large minibatch sgd: Training imagenet in 1 hour.
\newblock {\em arXiv:1706.02677}, 2017.

\bibitem{gu2022vector}
Shuyang Gu, Dong Chen, Jianmin Bao, Fang Wen, Bo Zhang, Dongdong Chen, Lu Yuan,
  and Baining Guo.
\newblock Vector quantized diffusion model for text-to-image synthesis.
\newblock In {\em CVPR}, pages 10696--10706, 2022.

\bibitem{He2016}
Kaiming He, Xiangyu Zhang, Shaoqing Ren, and Jian Sun.
\newblock Deep residual learning for image recognition.
\newblock In {\em CVPR}, 2016.

\bibitem{hendrycks2016gaussian}
Dan Hendrycks and Kevin Gimpel.
\newblock Gaussian error linear units (gelus).
\newblock {\em arXiv preprint arXiv:1606.08415}, 2016.

\bibitem{henighan2020scaling}
Tom Henighan, Jared Kaplan, Mor Katz, Mark Chen, Christopher Hesse, Jacob
  Jackson, Heewoo Jun, Tom~B Brown, Prafulla Dhariwal, Scott Gray, et~al.
\newblock Scaling laws for autoregressive generative modeling.
\newblock {\em arXiv preprint arXiv:2010.14701}, 2020.

\bibitem{heusel2017gans}
Martin Heusel, Hubert Ramsauer, Thomas Unterthiner, Bernhard Nessler, and Sepp
  Hochreiter.
\newblock Gans trained by a two time-scale update rule converge to a local nash
  equilibrium.
\newblock 2017.

\bibitem{ho2020ddpm}
Jonathan Ho, Ajay Jain, and Pieter Abbeel.
\newblock Denoising diffusion probabilistic models.
\newblock In {\em NeurIPS}, 2020.

\bibitem{ho2021cascaded}
Jonathan Ho, Chitwan Saharia, William Chan, David~J Fleet, Mohammad Norouzi,
  and Tim Salimans.
\newblock Cascaded diffusion models for high fidelity image generation.
\newblock {\em arXiv:2106.15282}, 2021.

\bibitem{ho2021classifier}
Jonathan Ho and Tim Salimans.
\newblock Classifier-free diffusion guidance.
\newblock In {\em NeurIPS 2021 Workshop on Deep Generative Models and
  Downstream Applications}, 2021.

\bibitem{hyvarinen2005estimation}
Aapo Hyv{\"a}rinen and Peter Dayan.
\newblock Estimation of non-normalized statistical models by score matching.
\newblock {\em Journal of Machine Learning Research}, 6(4), 2005.

\bibitem{isola2017image}
Phillip Isola, Jun-Yan Zhu, Tinghui Zhou, and Alexei~A Efros.
\newblock Image-to-image translation with conditional adversarial networks.
\newblock In {\em Proceedings of the IEEE conference on computer vision and
  pattern recognition}, pages 1125--1134, 2017.

\bibitem{jabri2022scalable}
Allan Jabri, David Fleet, and Ting Chen.
\newblock Scalable adaptive computation for iterative generation.
\newblock {\em arXiv preprint arXiv:2212.11972}, 2022.

\bibitem{janner2021trajectory}
Michael Janner, Qiyang Li, and Sergey Levine.
\newblock Offline reinforcement learning as one big sequence modeling problem.
\newblock In {\em NeurIPS}, 2021.

\bibitem{kaplan2020scaling}
Jared Kaplan, Sam McCandlish, Tom Henighan, Tom~B Brown, Benjamin Chess, Rewon
  Child, Scott Gray, Alec Radford, Jeffrey Wu, and Dario Amodei.
\newblock Scaling laws for neural language models.
\newblock {\em arXiv:2001.08361}, 2020.

\bibitem{Karras2022edm}
Tero Karras, Miika Aittala, Timo Aila, and Samuli Laine.
\newblock Elucidating the design space of diffusion-based generative models.
\newblock In {\em Proc. NeurIPS}, 2022.

\bibitem{karras2019style}
Tero Karras, Samuli Laine, and Timo Aila.
\newblock A style-based generator architecture for generative adversarial
  networks.
\newblock In {\em CVPR}, 2019.

\bibitem{kingma2014adam}
Diederik Kingma and Jimmy Ba.
\newblock Adam: A method for stochastic optimization.
\newblock In {\em ICLR}, 2015.

\bibitem{kingma2013auto}
Diederik~P Kingma and Max Welling.
\newblock Auto-encoding variational bayes.
\newblock {\em arXiv preprint arXiv:1312.6114}, 2013.

\bibitem{krizhevsky2012imagenet}
Alex Krizhevsky, Ilya Sutskever, and Geoffrey~E Hinton.
\newblock Imagenet classification with deep convolutional neural networks.
\newblock In {\em NeurIPS}, 2012.

\bibitem{kynkaanniemi2019improved}
Tuomas Kynk{\"a}{\"a}nniemi, Tero Karras, Samuli Laine, Jaakko Lehtinen, and
  Timo Aila.
\newblock Improved precision and recall metric for assessing generative models.
\newblock In {\em NeurIPS}, 2019.

\bibitem{loshchilov2017decoupled}
Ilya Loshchilov and Frank Hutter.
\newblock Decoupled weight decay regularization.
\newblock {\em arXiv:1711.05101}, 2017.

\bibitem{nash2021generating}
Charlie Nash, Jacob Menick, Sander Dieleman, and Peter~W Battaglia.
\newblock Generating images with sparse representations.
\newblock {\em arXiv preprint arXiv:2103.03841}, 2021.

\bibitem{nichol2021glide}
Alex Nichol, Prafulla Dhariwal, Aditya Ramesh, Pranav Shyam, Pamela Mishkin,
  Bob McGrew, Ilya Sutskever, and Mark Chen.
\newblock Glide: Towards photorealistic image generation and editing with
  text-guided diffusion models.
\newblock {\em arXiv:2112.10741}, 2021.

\bibitem{nichol2021improved}
Alexander~Quinn Nichol and Prafulla Dhariwal.
\newblock Improved denoising diffusion probabilistic models.
\newblock In {\em ICML}, 2021.

\bibitem{parmar2021cleanfid}
Gaurav Parmar, Richard Zhang, and Jun-Yan Zhu.
\newblock On aliased resizing and surprising subtleties in gan evaluation.
\newblock In {\em CVPR}, 2022.

\bibitem{parmar2018image}
Niki Parmar, Ashish Vaswani, Jakob Uszkoreit, Lukasz Kaiser, Noam Shazeer,
  Alexander Ku, and Dustin Tran.
\newblock Image transformer.
\newblock In {\em International conference on machine learning}, pages
  4055--4064. PMLR, 2018.

\bibitem{Peebles2022}
William Peebles, Ilija Radosavovic, Tim Brooks, Alexei Efros, and Jitendra
  Malik.
\newblock Learning to learn with generative models of neural network
  checkpoints.
\newblock {\em arXiv preprint arXiv:2209.12892}, 2022.

\bibitem{perez2018film}
Ethan Perez, Florian Strub, Harm De~Vries, Vincent Dumoulin, and Aaron
  Courville.
\newblock Film: Visual reasoning with a general conditioning layer.
\newblock In {\em AAAI}, 2018.

\bibitem{radford2021}
Alec Radford, Jong~Wook Kim, Chris Hallacy, Aditya Ramesh, Gabriel Goh,
  Sandhini Agarwal, Girish Sastry, Amanda Askell, Pamela Mishkin, Jack Clark,
  et~al.
\newblock Learning transferable visual models from natural language
  supervision.
\newblock In {\em ICML}, 2021.

\bibitem{Radford2018}
Alec Radford, Karthik Narasimhan, Tim Salimans, and Ilya Sutskever.
\newblock Improving language understanding by generative pre-training.
\newblock 2018.

\bibitem{Radford2019}
Alec Radford, Jeffrey Wu, Rewon Child, David Luan, Dario Amodei, Ilya
  Sutskever, et~al.
\newblock Language models are unsupervised multitask learners.
\newblock 2019.

\bibitem{radosavovic2019network}
Ilija Radosavovic, Justin Johnson, Saining Xie, Wan-Yen Lo, and Piotr
  Doll{\'a}r.
\newblock On network design spaces for visual recognition.
\newblock In {\em ICCV}, 2019.

\bibitem{radosavovic2020designing}
Ilija Radosavovic, Raj~Prateek Kosaraju, Ross Girshick, Kaiming He, and Piotr
  Doll{\'a}r.
\newblock Designing network design spaces.
\newblock In {\em CVPR}, 2020.

\bibitem{ramesh2022hierarchical}
Aditya Ramesh, Prafulla Dhariwal, Alex Nichol, Casey Chu, and Mark Chen.
\newblock Hierarchical text-conditional image generation with clip latents.
\newblock {\em arXiv:2204.06125}, 2022.

\bibitem{ramesh2021zero}
Aditya Ramesh, Mikhail Pavlov, Gabriel Goh, Scott Gray, Chelsea Voss, Alec
  Radford, Mark Chen, and Ilya Sutskever.
\newblock Zero-shot text-to-image generation.
\newblock In {\em ICML}, 2021.

\bibitem{rombach2021highresolution}
Robin Rombach, Andreas Blattmann, Dominik Lorenz, Patrick Esser, and Björn
  Ommer.
\newblock High-resolution image synthesis with latent diffusion models.
\newblock In {\em CVPR}, 2022.

\bibitem{ronneberger2015u}
Olaf Ronneberger, Philipp Fischer, and Thomas Brox.
\newblock U-net: Convolutional networks for biomedical image segmentation.
\newblock In {\em International Conference on Medical image computing and
  computer-assisted intervention}, pages 234--241. Springer, 2015.

\bibitem{imagen2022}
Chitwan Saharia, William Chan, Saurabh Saxena, Lala Li, Jay Whang, Emily
  Denton, Seyed Kamyar~Seyed Ghasemipour, Burcu~Karagol Ayan, S.~Sara Mahdavi,
  Rapha~Gontijo Lopes, Tim Salimans, Jonathan Ho, David~J Fleet, and Mohammad
  Norouzi.
\newblock Photorealistic text-to-image diffusion models with deep language
  understanding.
\newblock {\em arXiv:2205.11487}, 2022.

\bibitem{salimans2016gans}
Tim Salimans, Ian Goodfellow, Wojciech Zaremba, Vicki Cheung, Alec Radford, Xi
  Chen, and Xi Chen.
\newblock Improved techniques for training {GANs}.
\newblock In {\em NeurIPS}, 2016.

\bibitem{salimans2017pixelcnn++}
Tim Salimans, Andrej Karpathy, Xi Chen, and Diederik~P Kingma.
\newblock {PixelCNN++}: Improving the pixelcnn with discretized logistic
  mixture likelihood and other modifications.
\newblock {\em arXiv preprint arXiv:1701.05517}, 2017.

\bibitem{Sauer2021ARXIV}
Axel Sauer, Katja Schwarz, and Andreas Geiger.
\newblock Stylegan-xl: Scaling stylegan to large diverse datasets.
\newblock In {\em SIGGRAPH}, 2022.

\bibitem{sohl2015thermodynamics}
Jascha Sohl-Dickstein, Eric Weiss, Niru Maheswaranathan, and Surya Ganguli.
\newblock Deep unsupervised learning using nonequilibrium thermodynamics.
\newblock In {\em ICML}, 2015.

\bibitem{song2020denoising}
Jiaming Song, Chenlin Meng, and Stefano Ermon.
\newblock Denoising diffusion implicit models.
\newblock {\em arXiv:2010.02502}, 2020.

\bibitem{song2019generative}
Yang Song and Stefano Ermon.
\newblock Generative modeling by estimating gradients of the data distribution.
\newblock In {\em NeurIPS}, 2019.

\bibitem{steiner2021train}
Andreas Steiner, Alexander Kolesnikov, Xiaohua Zhai, Ross Wightman, Jakob
  Uszkoreit, and Lucas Beyer.
\newblock How to train your {ViT}? data, augmentation, and regularization in
  vision transformers.
\newblock {\em TMLR}, 2022.

\bibitem{van2016conditional}
Aaron Van~den Oord, Nal Kalchbrenner, Lasse Espeholt, Oriol Vinyals, Alex
  Graves, et~al.
\newblock Conditional image generation with pixelcnn decoders.
\newblock {\em Advances in neural information processing systems}, 29, 2016.

\bibitem{van2017neural}
Aaron Van Den~Oord, Oriol Vinyals, et~al.
\newblock Neural discrete representation learning.
\newblock {\em Advances in neural information processing systems}, 30, 2017.

\bibitem{Vaswani2017}
Ashish Vaswani, Noam Shazeer, Niki Parmar, Jakob Uszkoreit, Llion Jones,
  Aidan~N Gomez, {\L}ukasz Kaiser, and Illia Polosukhin.
\newblock Attention is all you need.
\newblock In {\em NeurIPS}, 2017.

\bibitem{Xiao2021}
Tete Xiao, Piotr Dollar, Mannat Singh, Eric Mintun, Trevor Darrell, and Ross
  Girshick.
\newblock Early convolutions help transformers see better.
\newblock In {\em NeurIPS}, 2021.

\bibitem{yu2022scaling}
Jiahui Yu, Yuanzhong Xu, Jing~Yu Koh, Thang Luong, Gunjan Baid, Zirui Wang,
  Vijay Vasudevan, Alexander Ku, Yinfei Yang, Burcu~Karagol Ayan, et~al.
\newblock Scaling autoregressive models for content-rich text-to-image
  generation.
\newblock {\em arXiv:2206.10789}, 2022.

\bibitem{zhai2022scaling}
Xiaohua Zhai, Alexander Kolesnikov, Neil Houlsby, and Lucas Beyer.
\newblock Scaling vision transformers.
\newblock In {\em CVPR}, 2022.

\end{thebibliography}
}
\clearpage

\appendix
\twocolumn[{
\renewcommand\twocolumn[1][]{#1}%
\begin{center}
    \includegraphics[width=\textwidth]{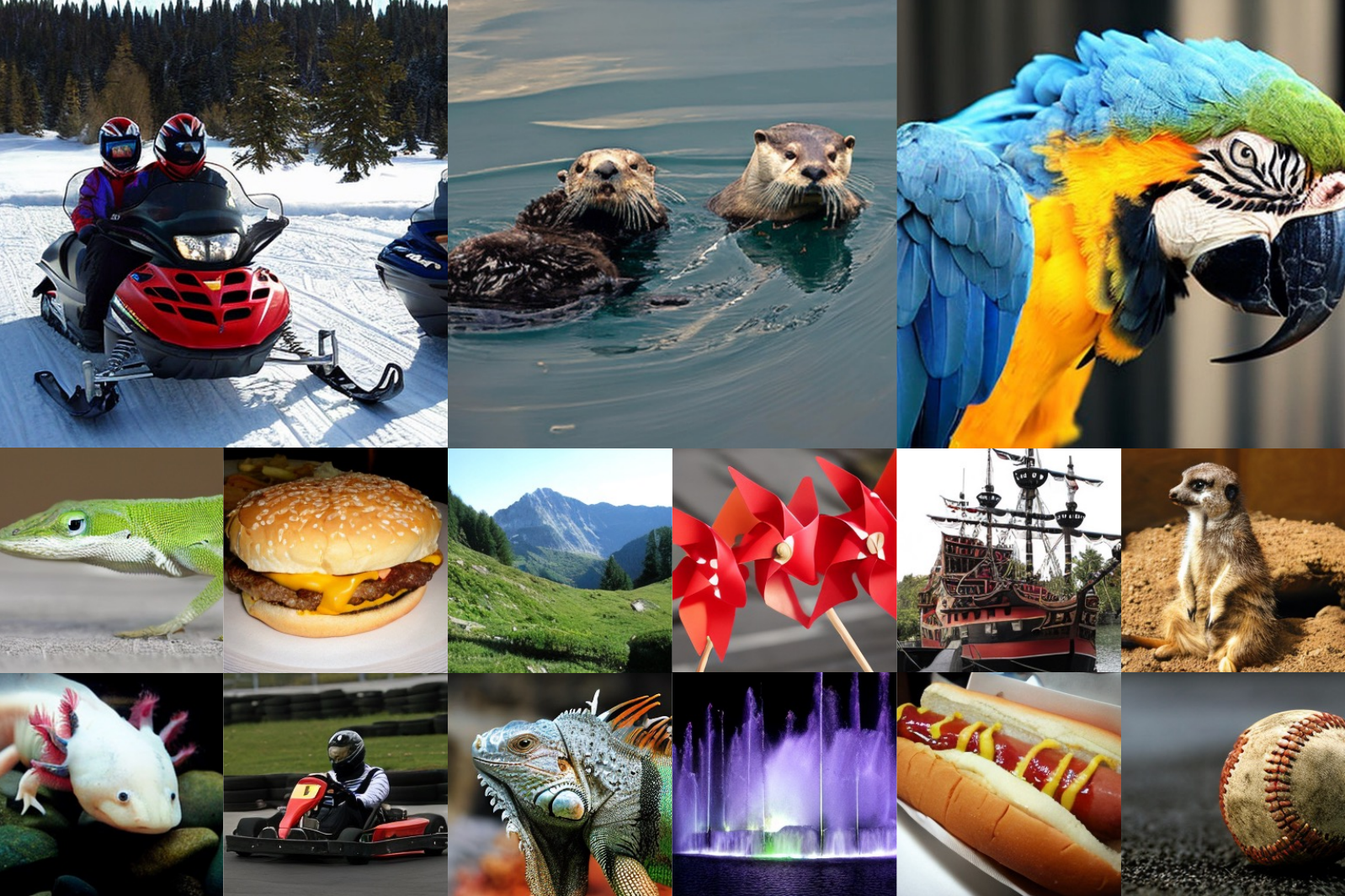}
    \captionof{figure}{\textbf{Additional selected samples from our 512$\times$512 and 256$\times$256 resolution DiT-XL/2 models.} We use a classifier-free guidance scale of 6.0 for the $512\times512$ model and 4.0 for the $256\times256$ model. Both models use the ft-EMA VAE decoder.}\label{fig:sota512}
\end{center}
}]

\begin{table*}
\centering
\small
\scalebox{0.8}{
\begin{tabular}{l c c c c c c c c}
\toprule
Model         & Image Resolution   & Flops (G) & Params (M) & Training Steps (K) & Batch Size & Learning Rate & DiT Block & FID-50K (no guidance)  \\
\midrule 
DiT-S/8   &   $256\times256$   &     0.36   &   33       &  400 &  256 & $1 \times 10^{-4}$ & adaLN-Zero & 153.60 \\
DiT-S/4   &   $256\times256$   &     1.41   &   33       &  400 &  256 & $1 \times 10^{-4}$ & adaLN-Zero & 100.41 \\
DiT-S/2   &   $256\times256$   &     6.06   &   33       &  400 &  256 & $1 \times 10^{-4}$ & adaLN-Zero & 68.40 \\
\midrule
DiT-B/8   &   $256\times256$   &     1.42    &   131    &  400 &  256 & $1 \times 10^{-4}$ & adaLN-Zero & 122.74 \\
DiT-B/4   &   $256\times256$   &     5.56    &   130    &  400 &  256 & $1 \times 10^{-4}$ & adaLN-Zero & 68.38 \\
DiT-B/2   &   $256\times256$   &     23.01    &   130    &  400 &  256 & $1 \times 10^{-4}$ & adaLN-Zero & 43.47 \\
\midrule
DiT-L/8  &    $256\times256$   &     5.01    &   459   &  400 &  256 & $1 \times 10^{-4}$ & adaLN-Zero & 118.87 \\
DiT-L/4  &    $256\times256$   &     19.70    &   458   &  400 &  256 & $1 \times 10^{-4}$ & adaLN-Zero & 45.64 \\
DiT-L/2  &    $256\times256$   &     80.71    &   458   &  400 &  256 & $1 \times 10^{-4}$ & adaLN-Zero & 23.33 \\
\midrule
DiT-XL/8 &    $256\times256$  &      7.39     &   676  &  400 &  256 & $1 \times 10^{-4}$ & adaLN-Zero & 106.41 \\
DiT-XL/4 &    $256\times256$  &      29.05     &   675  &  400 &  256 & $1 \times 10^{-4}$ & adaLN-Zero & 43.01 \\
DiT-XL/2 &    $256\times256$  &      118.64     &   675  &  400 &  256 & $1 \times 10^{-4}$ & adaLN-Zero & 19.47 \\
DiT-XL/2 &    $256\times256$  &      119.37     &   449  &  400 &  256 & $1 \times 10^{-4}$ & in-context & 35.24 \\
DiT-XL/2 &    $256\times256$  &      137.62     &   598  &  400 &  256 & $1 \times 10^{-4}$ & cross-attention & 26.14 \\
DiT-XL/2 &    $256\times256$  &      118.56     &   600  &  400 &  256 & $1 \times 10^{-4}$ & adaLN & 25.21 \\
DiT-XL/2 &    $256\times256$  &      118.64     &   675  &  2352&  256 & $1 \times 10^{-4}$ & adaLN-Zero & 10.67 \\
DiT-XL/2 &    $256\times256$  &      118.64     &   675  &  7000&  256 & $1 \times 10^{-4}$ & adaLN-Zero & 9.62 \\
\midrule
DiT-XL/2 &    $512\times512$  &      524.60     &   675  &  1301&  256 & $1 \times 10^{-4}$ & adaLN-Zero & 13.78 \\
DiT-XL/2 &    $512\times512$  &      524.60     &   675  &  3000&  256 & $1 \times 10^{-4}$ & adaLN-Zero & 11.93 \\
\bottomrule
\end{tabular}}
\caption{\textbf{Details of all DiT models.} We report detailed information about every DiT model in our paper. Note that FID-50K here is computed \textit{without} classifier-free guidance. Parameter and flop counts exclude the VAE model which contains 84M parameters across the encoder and decoder. For both the $256\times256$ and $512\times512$ DiT-XL/2 models, we never observed FID saturate and continued training them as long as possible. Numbers reported in this table use the ft-MSE VAE decoder.}
\label{tbl:model_stats}
\end{table*}

\section{Additional Implementation Details}
\begin{figure*}\centering
\includegraphics[width=0.95\linewidth]{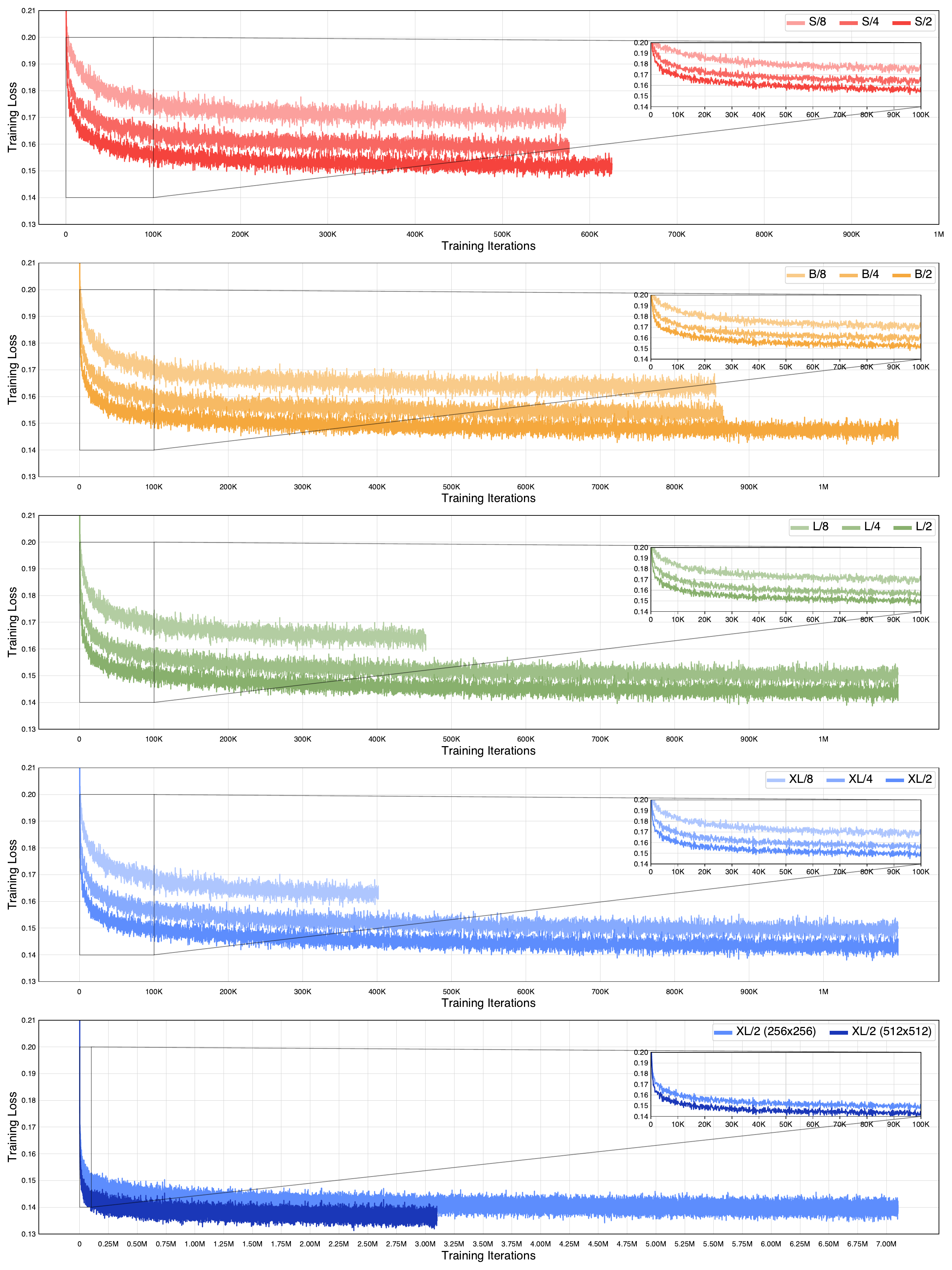}
\vspace{-4mm}
\caption{\textbf{Training loss curves for all DiT models.} We plot the loss over training for all DiT models (the sum of the noise prediction mean-squared error and $\mathcal{D}_{KL}$). We also highlight early training behavior. Note that scaled-up DiT models exhibit lower training losses.}
\label{fig:losses}
\end{figure*}

We include detailed information about all of our DiT models in Table~\ref{tbl:model_stats}, including both $256\times256$ and $512\times512$ models. In Figure~\ref{fig:losses}, we report DiT training loss curves. Finally, we also include Gflop counts for DDPM U-Net models from ADM and LDM in Table~\ref{tbl:baseline-flops}.
\vspace{-3mm}
\paragraph{DiT model details.} To embed input timesteps, we use a 256-dimensional frequency embedding~\cite{dhariwal2021adm} followed by a two-layer MLP with dimensionality equal to the transformer's hidden size and SiLU activations. Each adaLN layer feeds the sum of the timestep and class embeddings into a SiLU nonlinearity and a linear layer with output neurons equal to either $4\times$ (adaLN) or $6\times$ (adaLN-Zero) the transformer's hidden size. We use GELU nonlinearities (approximated with tanh) in the core transformer ~\cite{hendrycks2016gaussian}.
\vspace{-3mm}
\paragraph{Classifier-free guidance on a subset of channels.} In our experiments using classifier-free guidance, we applied guidance only to the first three channels of the latents instead of all four channels. Upon investigating, we found that three-channel guidance and four-channel guidance give similar results (in terms of FID) when simply adjusting the scale factor. Specifically, three-channel guidance with a scale of $(1 + x)$ appears reasonably well-approximated by four-channel guidance with a scale of $(1 + \frac{3}{4}x)$ (e.g., three-channel guidance with a scale of $1.5$ gives an FID-50K of 2.27, and four-channel guidance with a scale of $1.375$ gives an FID-50K of 2.20). It is somewhat interesting that applying guidance to a subset of elements can still yield good performance, and we leave it to future work to explore this phenomenon further.

\begin{figure*}\centering
\includegraphics[width=\linewidth]{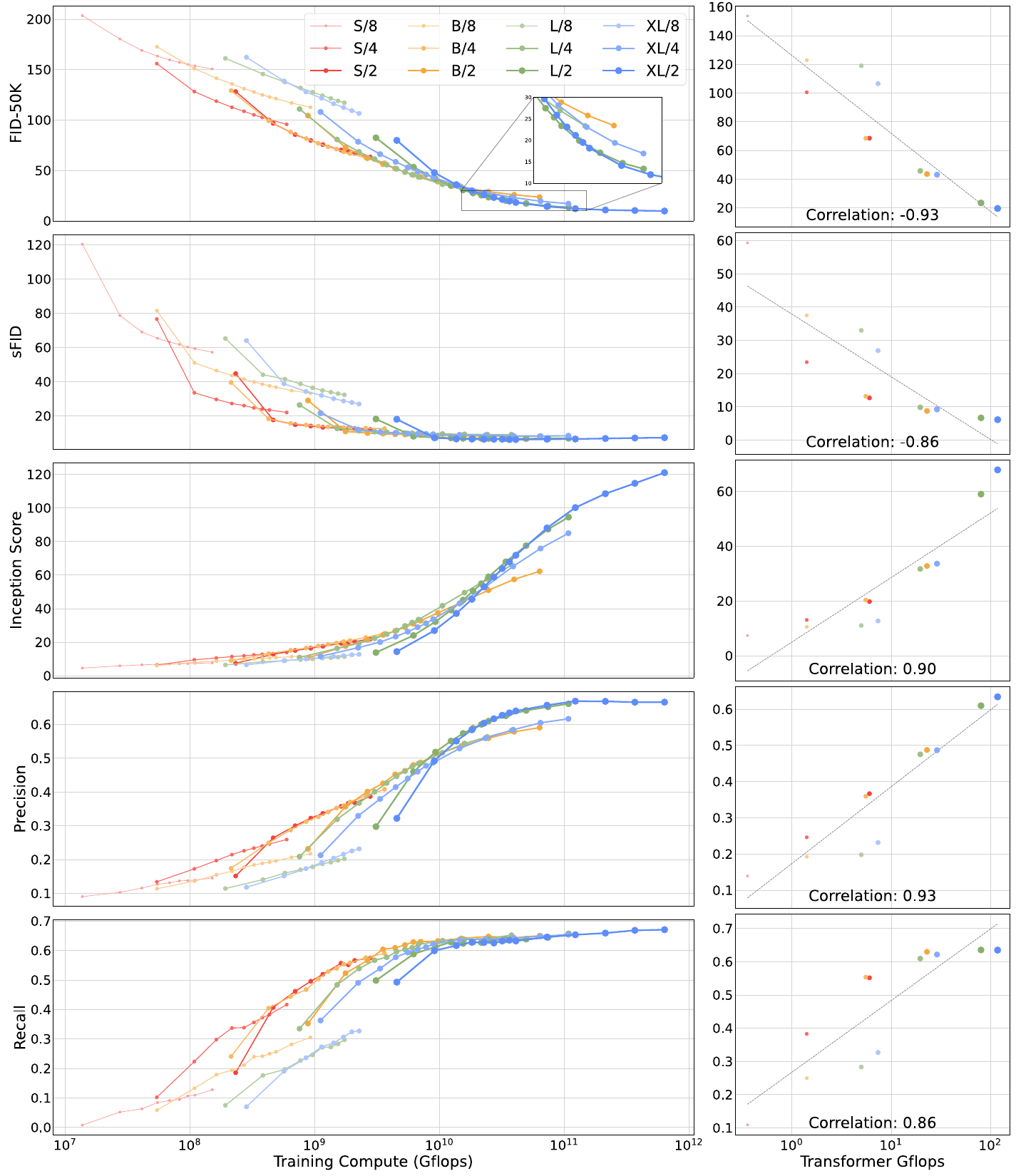}
\vspace{-4mm}
\caption{\textbf{DiT scaling behavior on several generative modeling metrics.} \emph{Left:} We plot model performance as a function of total training compute for FID, sFID, Inception Score, Precision and Recall. \emph{Right:} We plot model performance at 400K training steps for all 12 DiT variants against transformer Gflops, finding strong correlations across metrics. All values were computed using the ft-MSE VAE decoder.}
\label{fig:training_complexity_ISPR}
\end{figure*}

\section{Model Samples}

\begin{figure}\centering
\includegraphics[width=\linewidth]{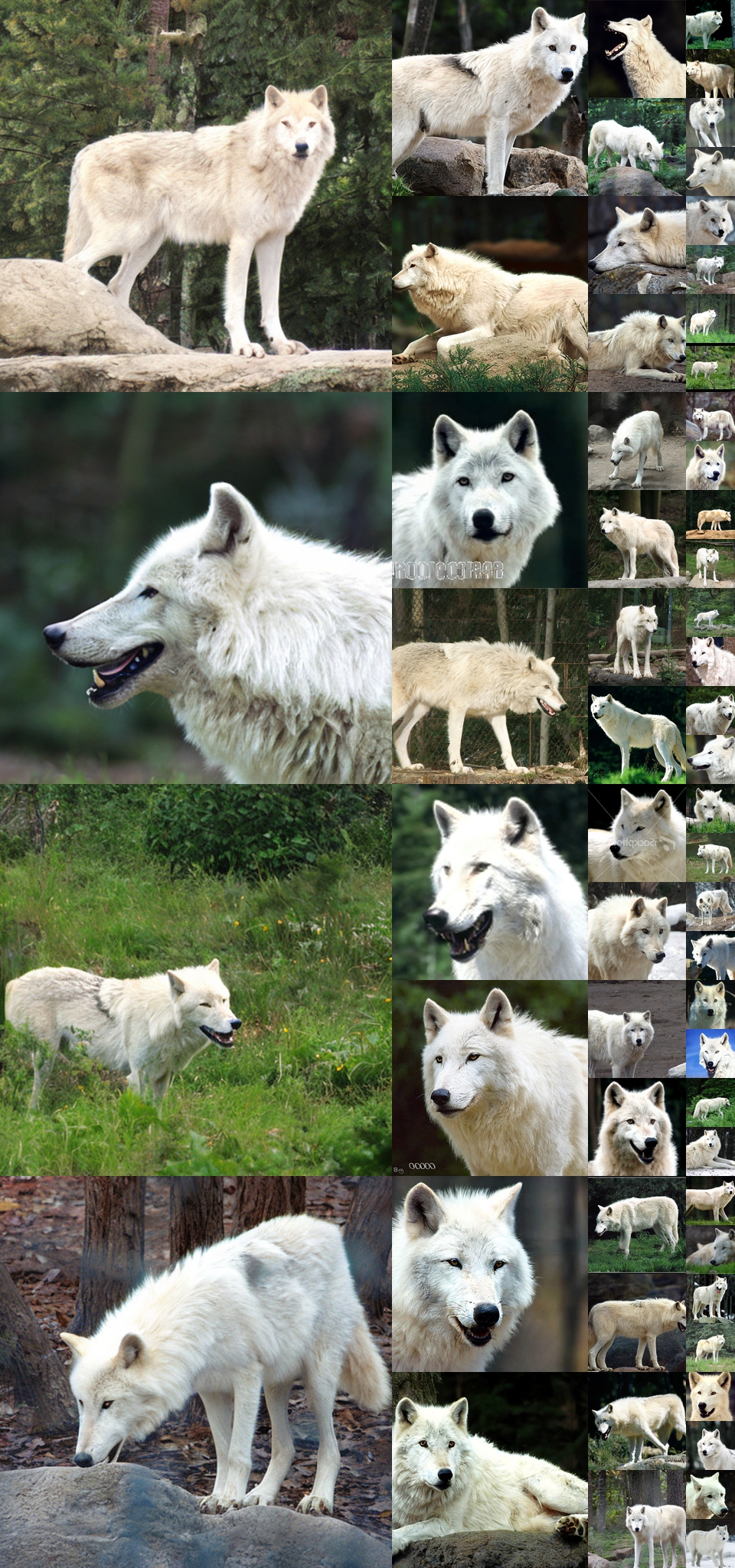}
\caption{\textbf{Uncurated $512\times512$ DiT-XL/2 samples.} \\Classifier-free guidance scale = 4.0\\Class label = ``arctic wolf" (270)}\vspace{-2mm}
\label{fig:samples512_1}
\end{figure}

\begin{figure}\centering
\includegraphics[width=\linewidth]{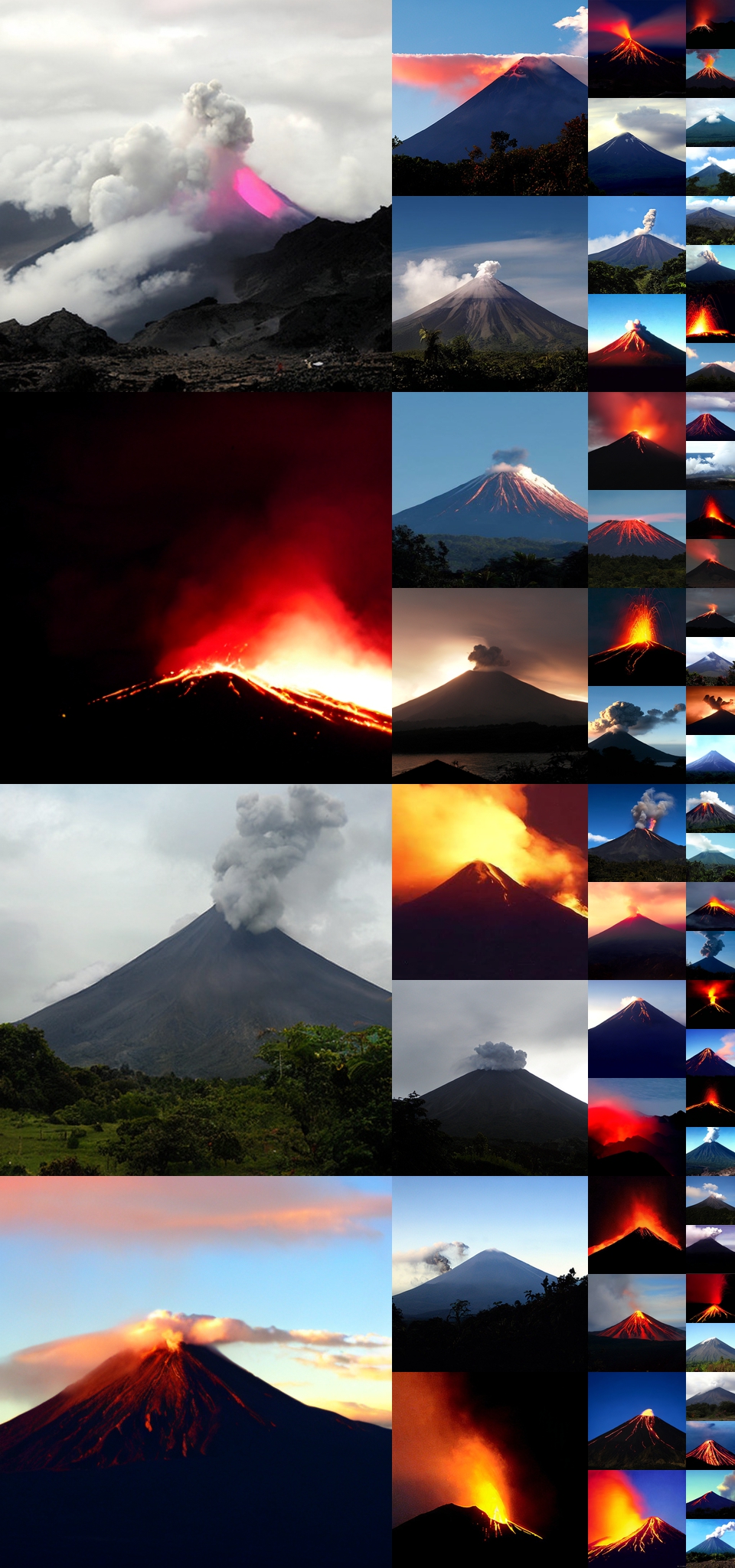}
\caption{\textbf{Uncurated $512\times512$ DiT-XL/2 samples.} \\Classifier-free guidance scale = 4.0\\Class label = ``volcano" (980)}\vspace{-2mm}
\label{fig:samples512_2}
\end{figure}

\begin{figure}\centering
\includegraphics[width=\linewidth]{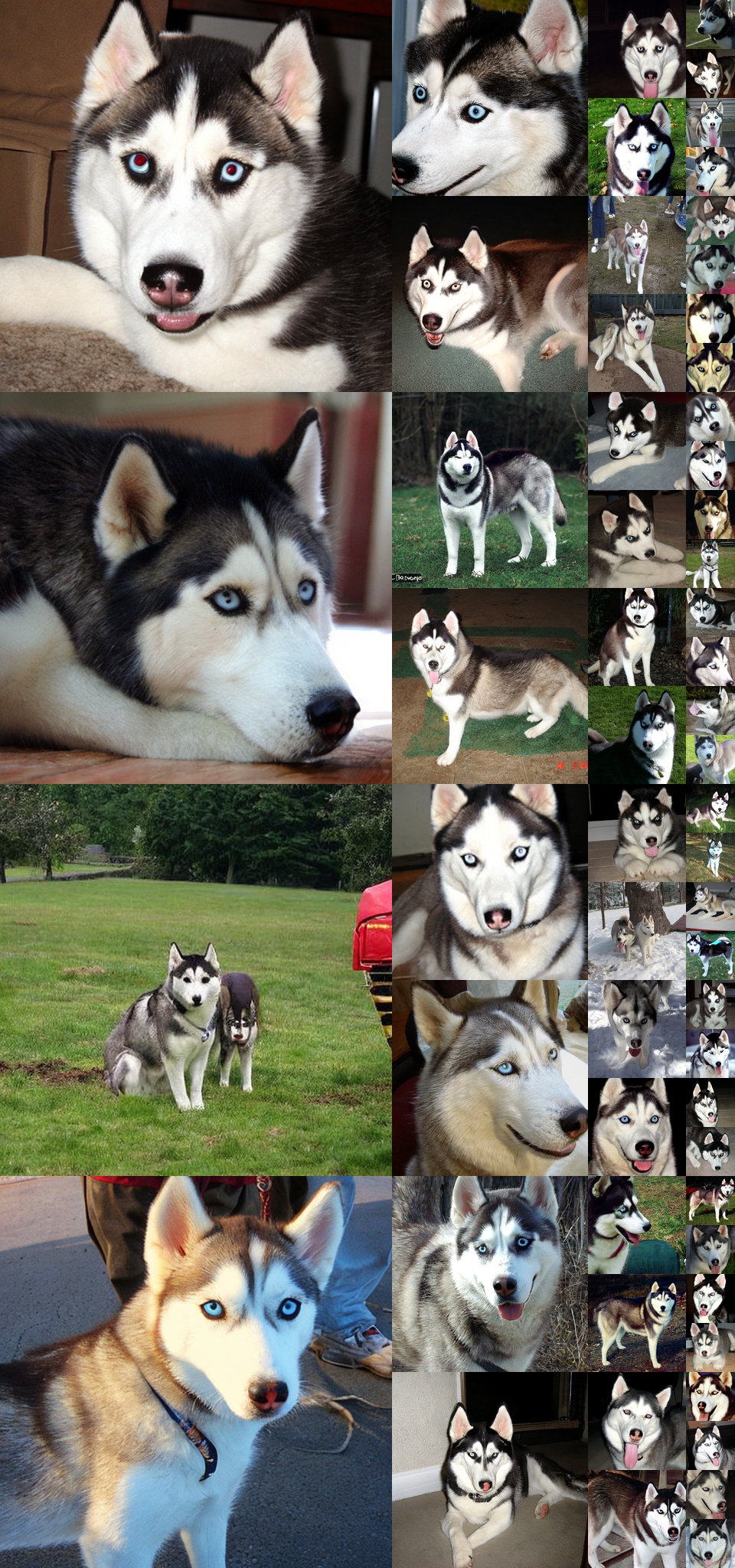}
\caption{\textbf{Uncurated $512\times512$ DiT-XL/2 samples.} \\Classifier-free guidance scale = 4.0\\Class label = ``husky" (250)}\vspace{-2mm}
\label{fig:samples512_3}
\end{figure}

\begin{figure}\centering
\includegraphics[width=\linewidth]{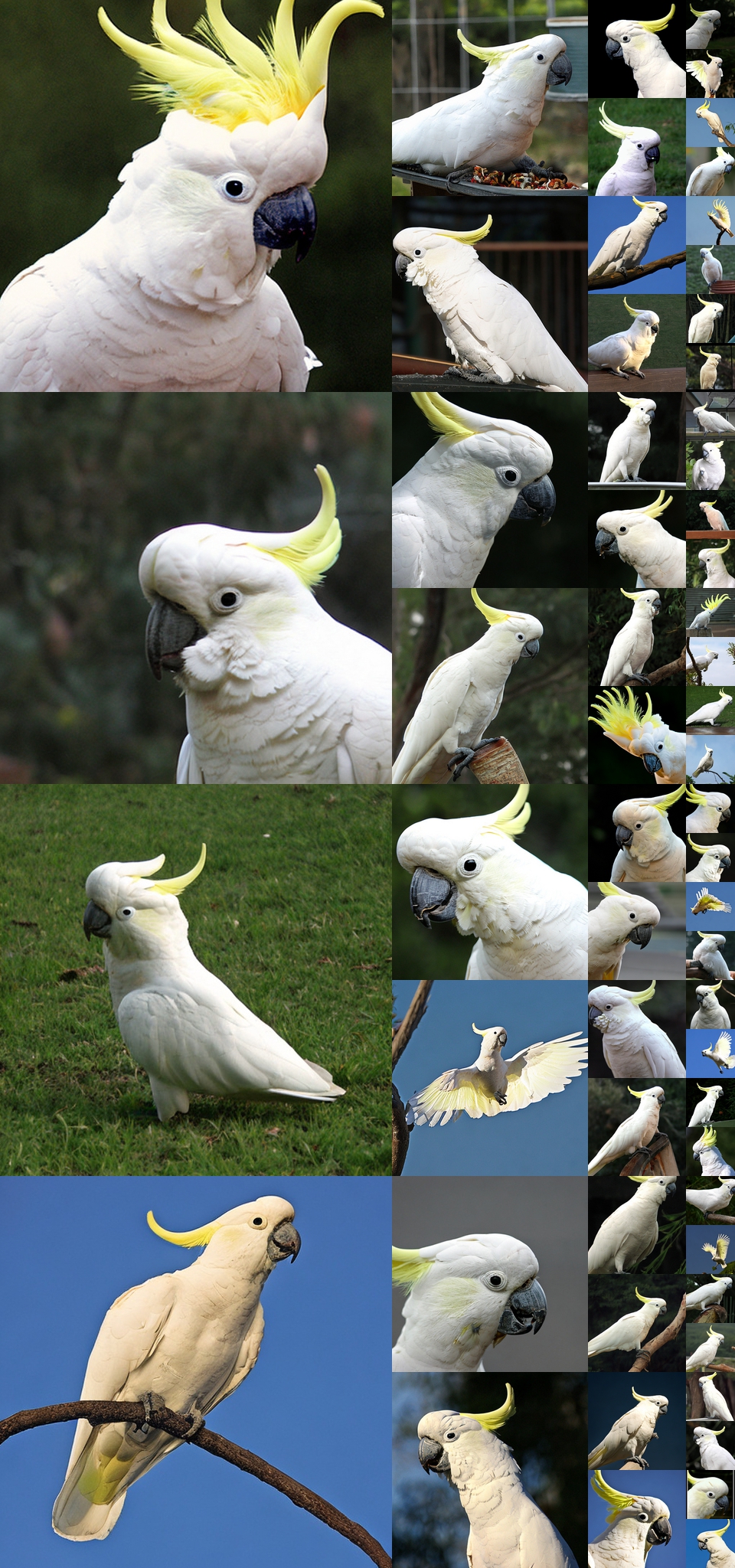}
\caption{\textbf{Uncurated $512\times512$ DiT-XL/2 samples.} \\Classifier-free guidance scale = 4.0\\Class label = ``sulphur-crested cockatoo" (89)}\vspace{-2mm}
\label{fig:samples512_4}
\end{figure}

\begin{figure}\centering
\includegraphics[width=\linewidth]{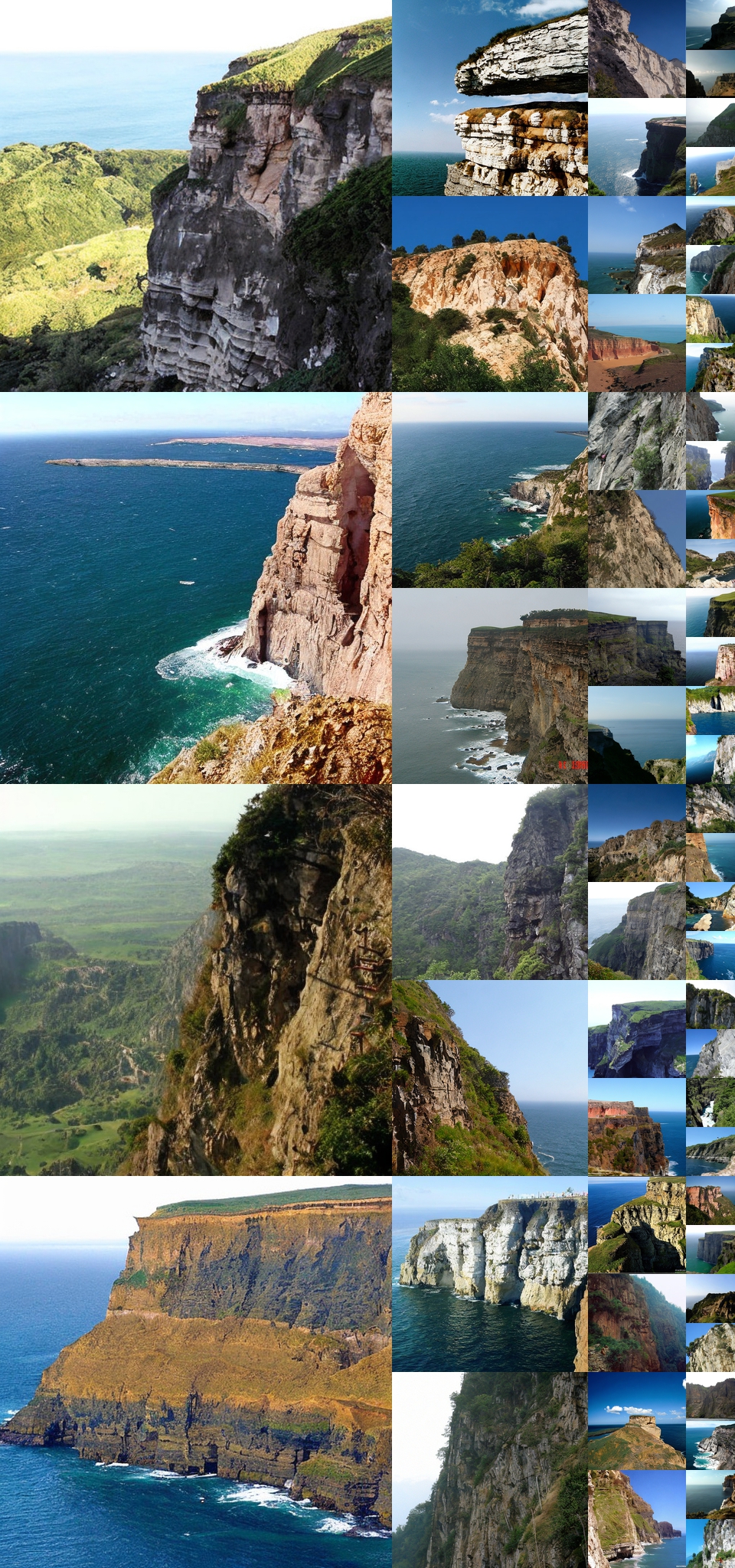}
\caption{\textbf{Uncurated $512\times512$ DiT-XL/2 samples.} \\Classifier-free guidance scale = 4.0\\Class label = ``cliff drop-off" (972)}\vspace{-2mm}
\label{fig:samples512_5}
\end{figure}

\begin{figure}\centering
\includegraphics[width=\linewidth]{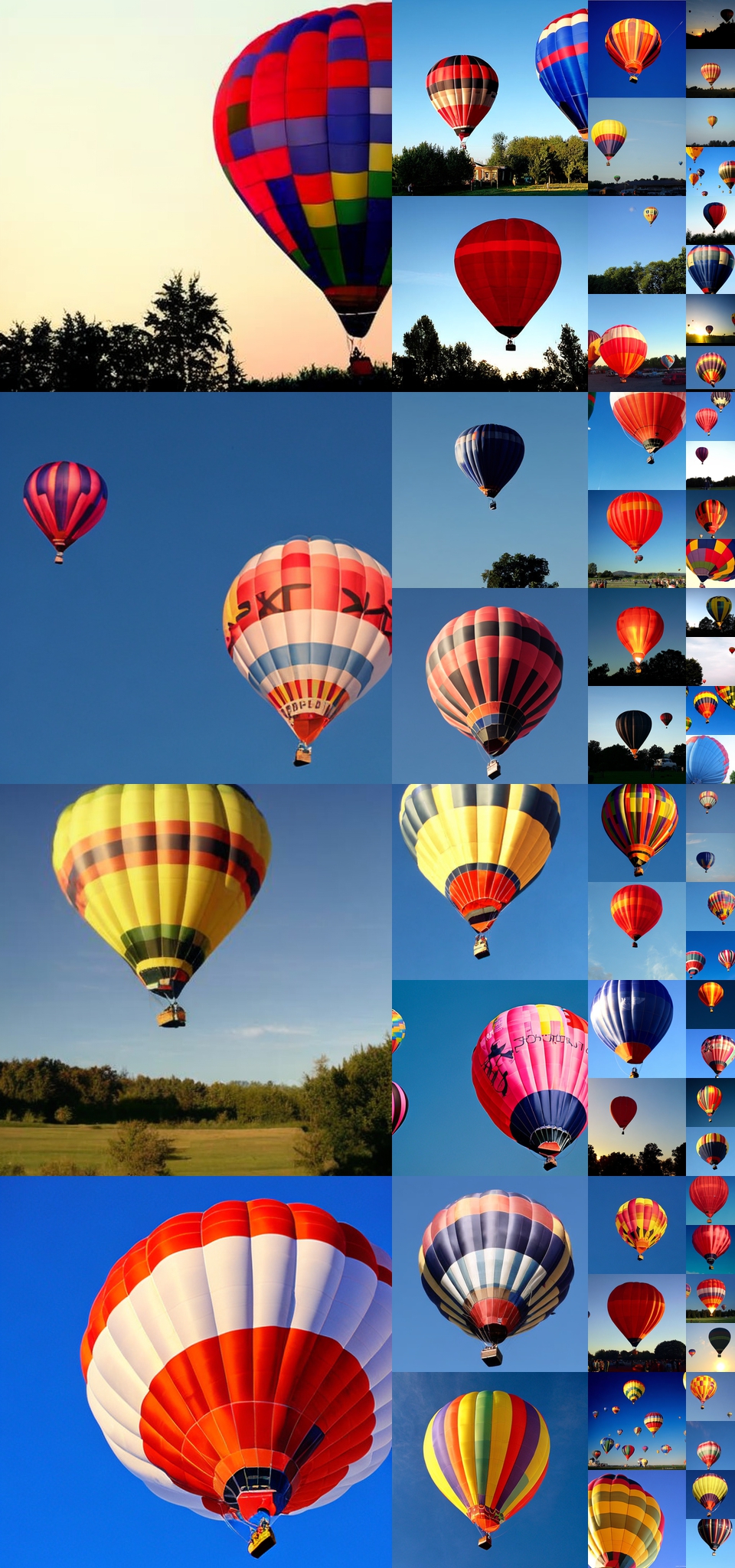}
\caption{\textbf{Uncurated $512\times512$ DiT-XL/2 samples.} \\Classifier-free guidance scale = 4.0\\Class label = ``balloon" (417)}\vspace{-2mm}
\label{fig:samples512_6}
\end{figure}

\begin{figure}\centering
\includegraphics[width=\linewidth]{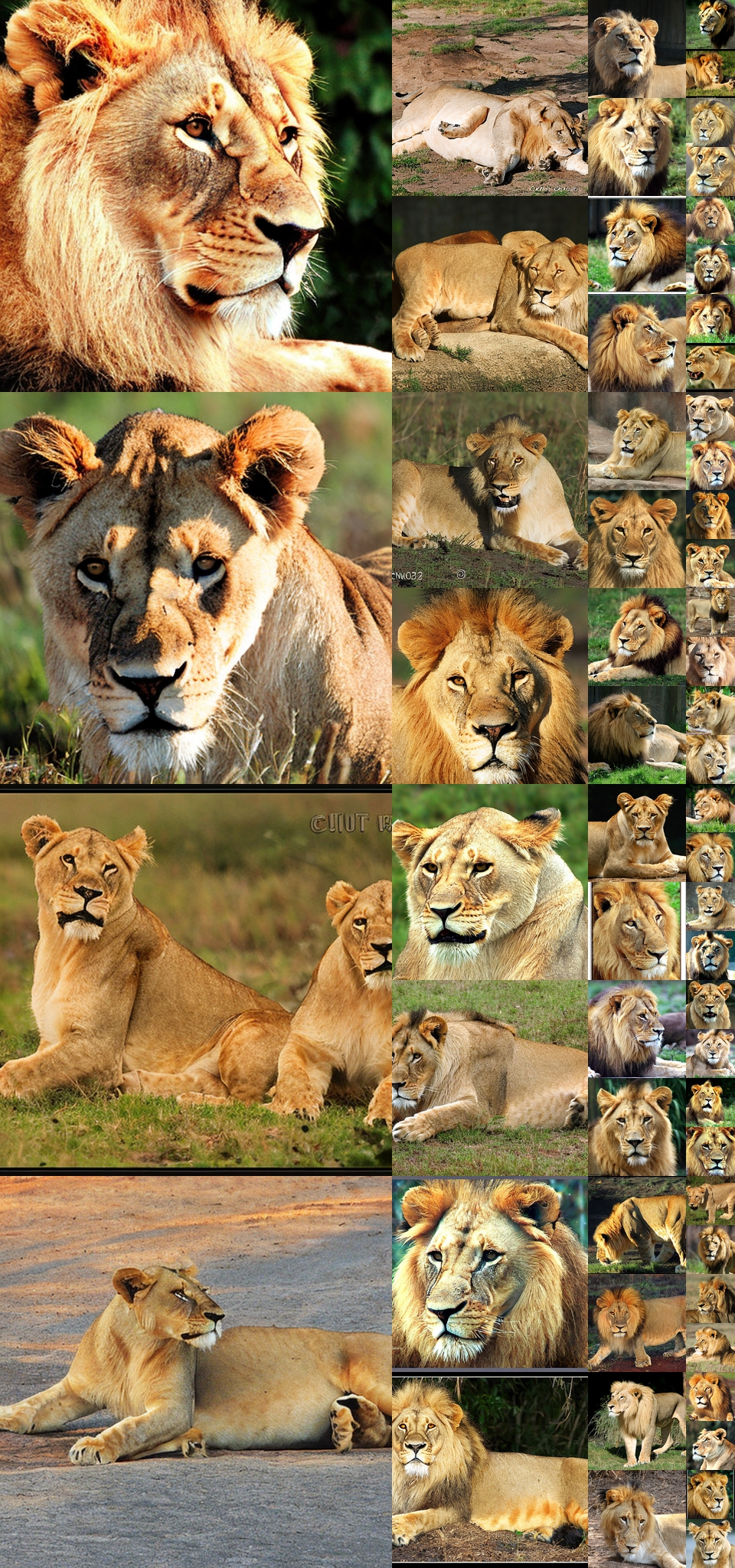}
\caption{\textbf{Uncurated $512\times512$ DiT-XL/2 samples.} \\Classifier-free guidance scale = 4.0\\Class label = ``lion" (291)}\vspace{-2mm}
\label{fig:samples512_7}
\end{figure}

\begin{figure}\centering
\includegraphics[width=\linewidth]{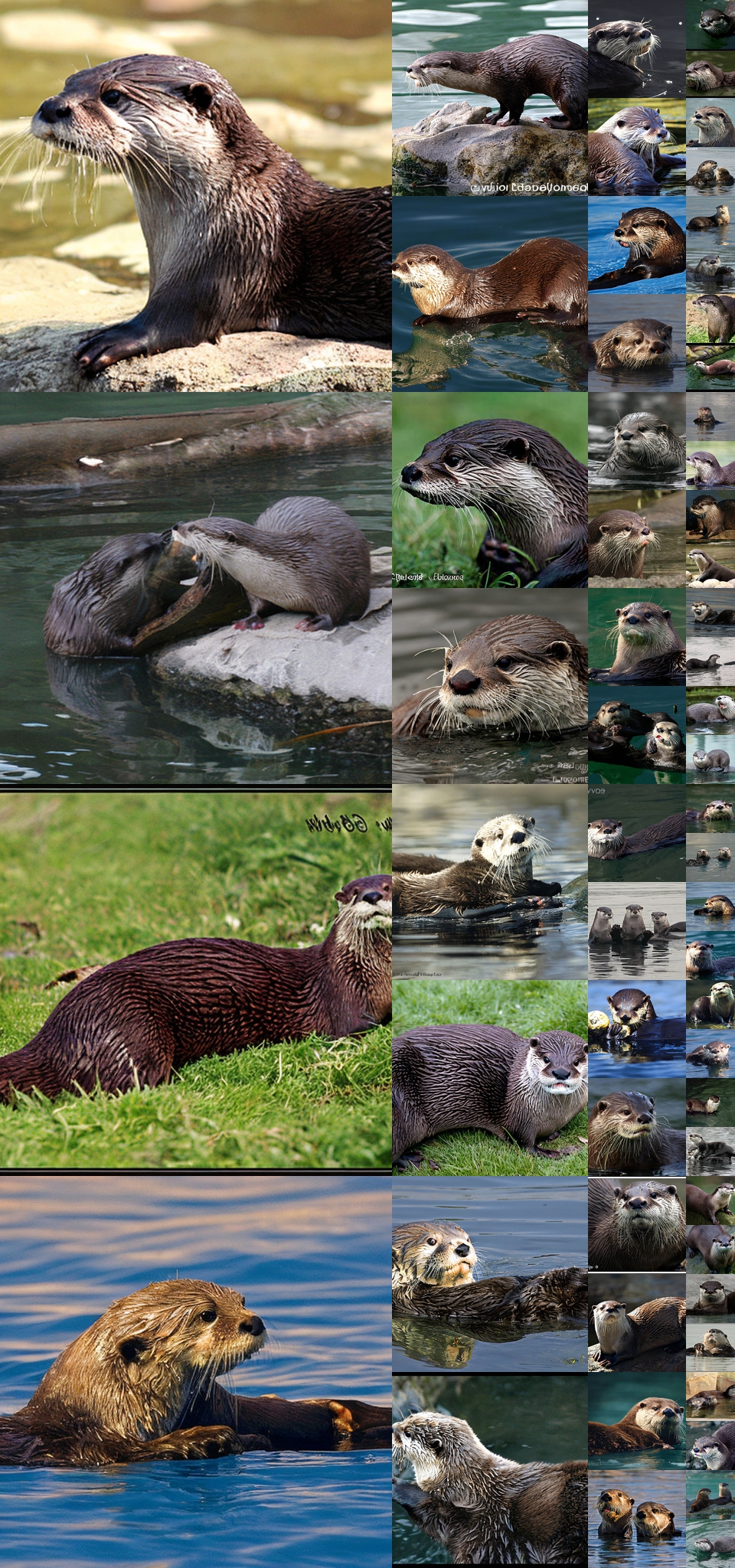}
\caption{\textbf{Uncurated $512\times512$ DiT-XL/2 samples.} \\Classifier-free guidance scale = 4.0\\Class label = ``otter" (360)}\vspace{-2mm}
\label{fig:samples512_8}
\end{figure}

\begin{figure}\centering
\includegraphics[width=\linewidth]{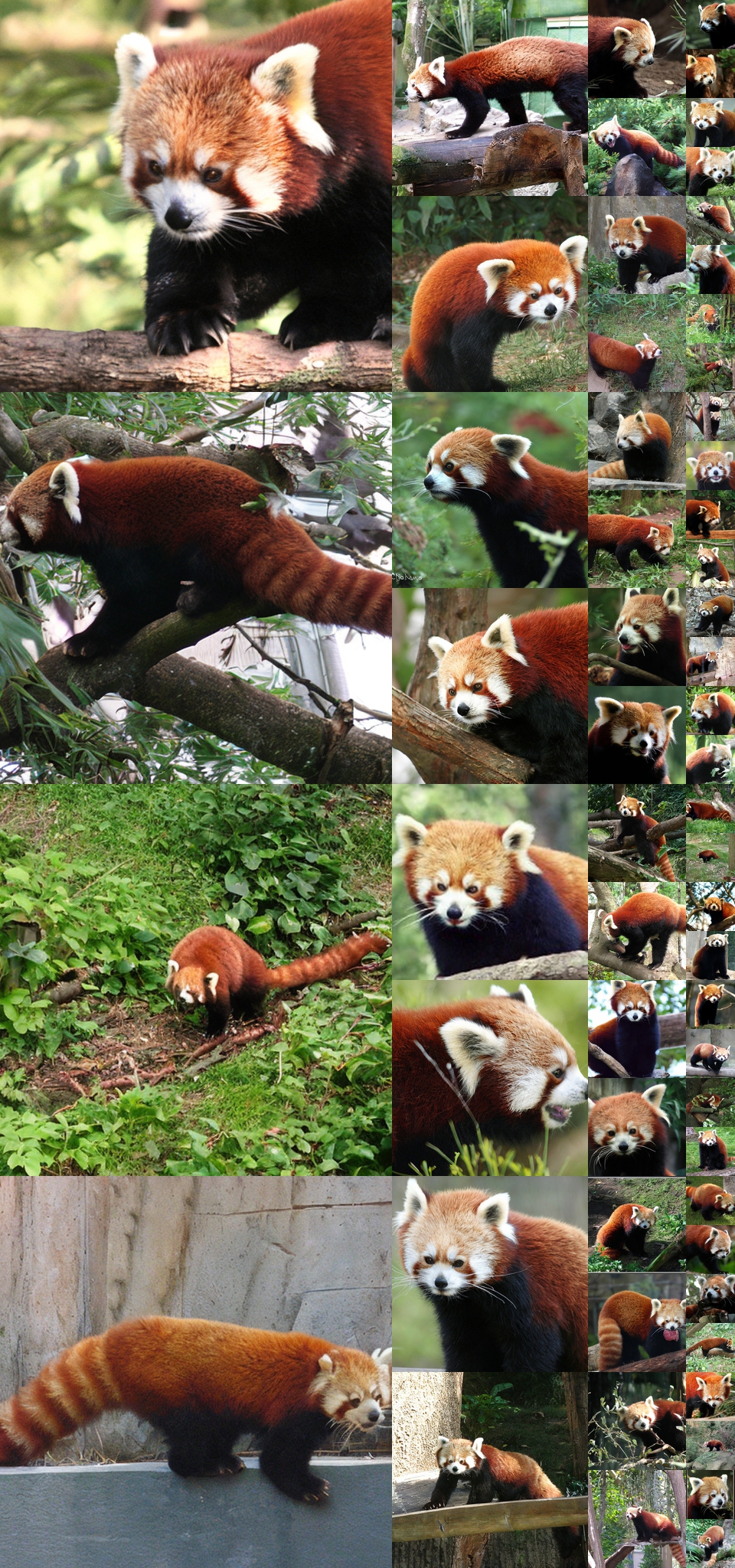}
\caption{\textbf{Uncurated $512\times512$ DiT-XL/2 samples.} \\Classifier-free guidance scale = 2.0\\Class label = ``red panda" (387)}\vspace{-2mm}
\label{fig:samples512_9}
\end{figure}

\begin{figure}\centering
\includegraphics[width=\linewidth]{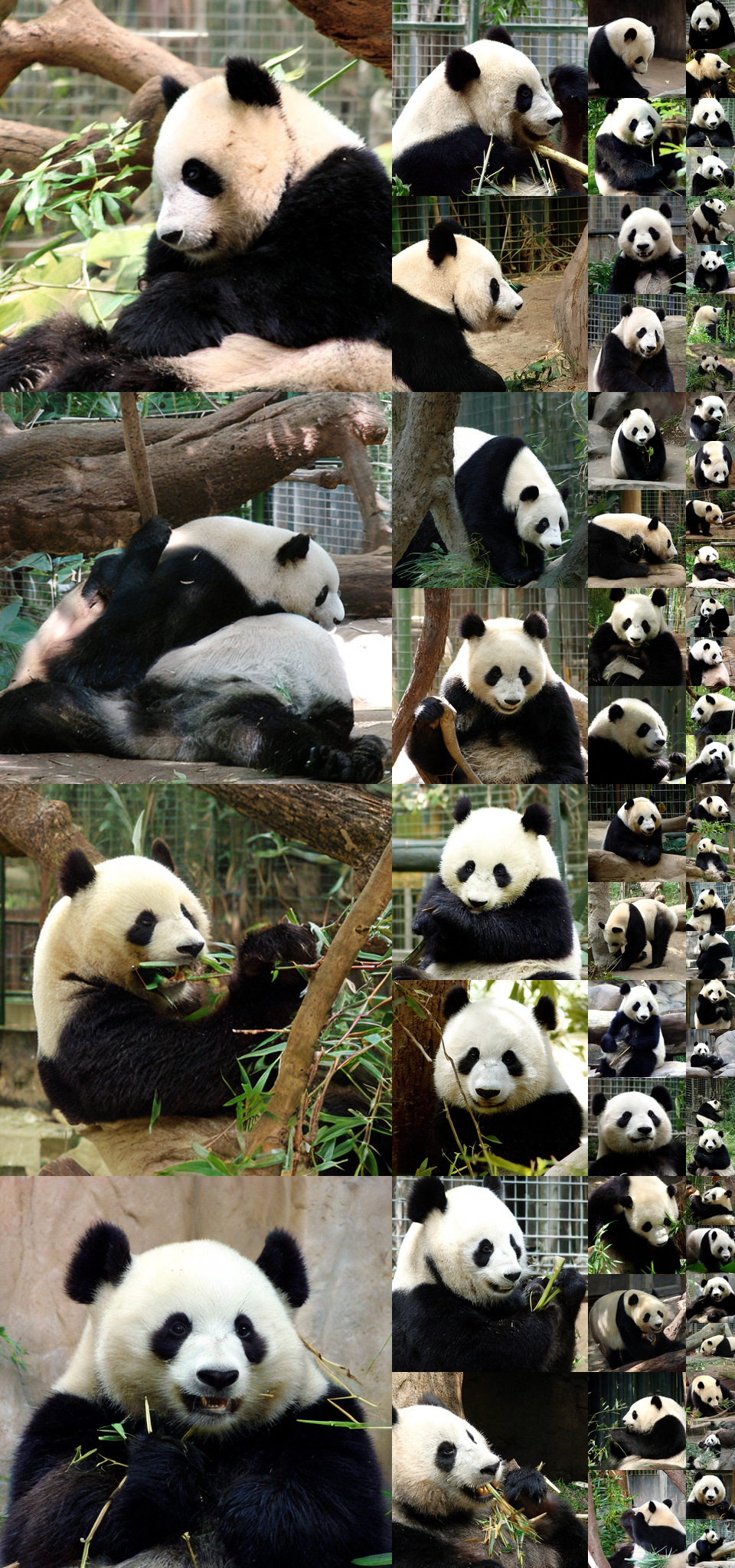}
\caption{\textbf{Uncurated $512\times512$ DiT-XL/2 samples.} \\Classifier-free guidance scale = 2.0\\Class label = ``panda" (388)}\vspace{-2mm}
\label{fig:samples512_10}
\end{figure}

\begin{figure}\centering
\includegraphics[width=\linewidth]{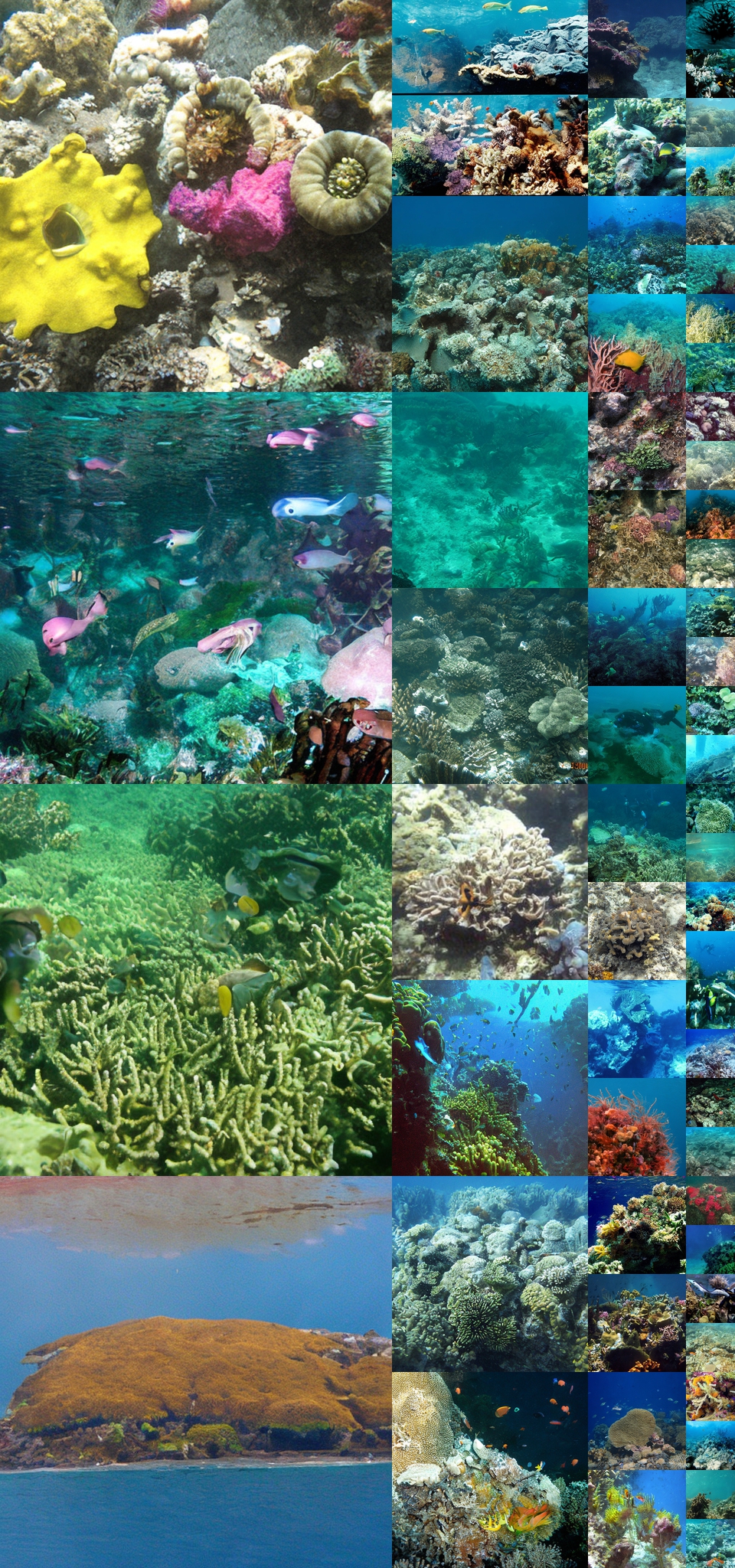}
\caption{\textbf{Uncurated $512\times512$ DiT-XL/2 samples.} \\Classifier-free guidance scale = 1.5\\Class label = ``coral reef" (973)}\vspace{-2mm}
\label{fig:samples512_11}
\end{figure}

\begin{figure}\centering
\includegraphics[width=\linewidth]{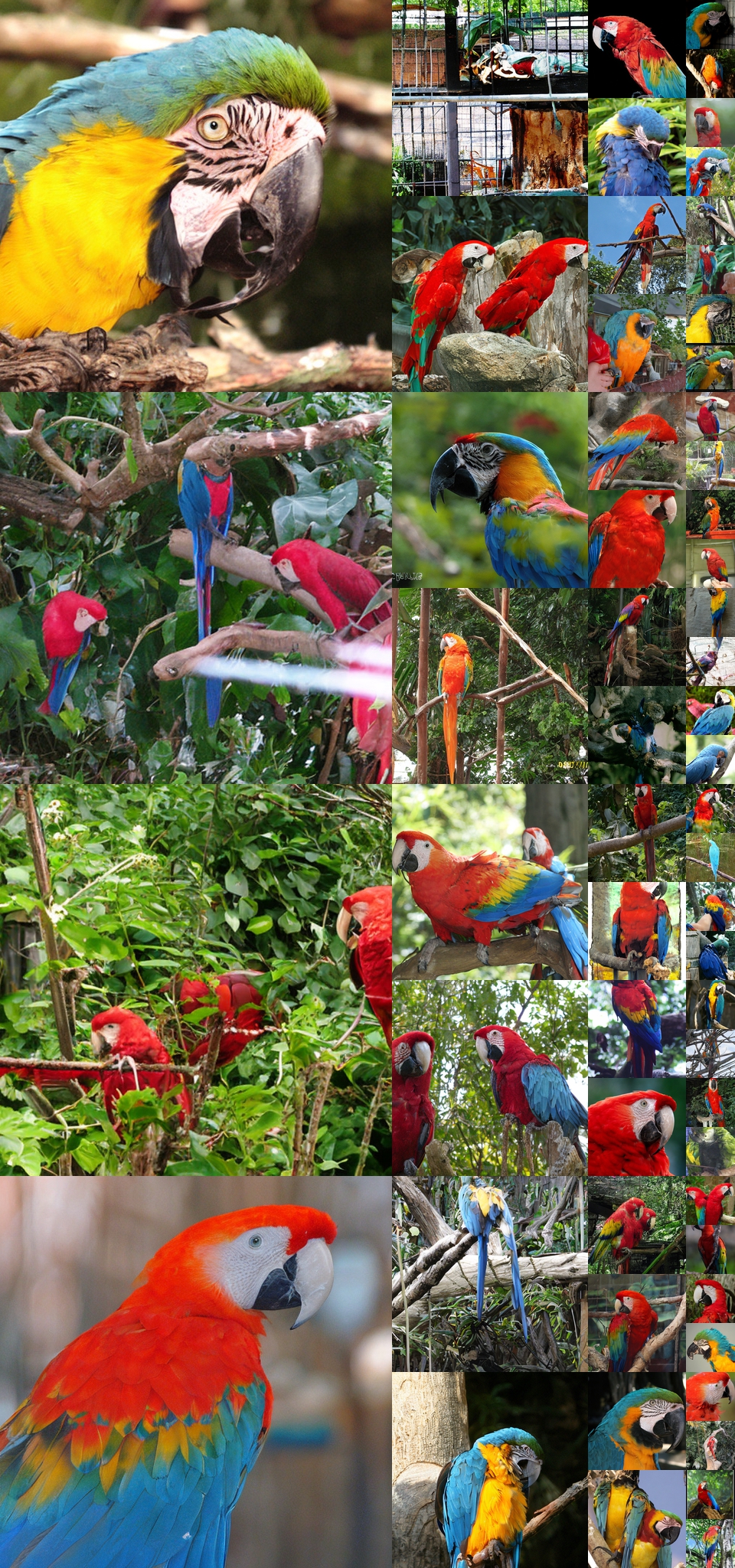}
\caption{\textbf{Uncurated $512\times512$ DiT-XL/2 samples.} \\Classifier-free guidance scale = 1.5\\Class label = ``macaw" (88)}\vspace{-2mm}
\label{fig:samples512_12}
\end{figure}

\begin{figure}\centering
\includegraphics[width=\linewidth]{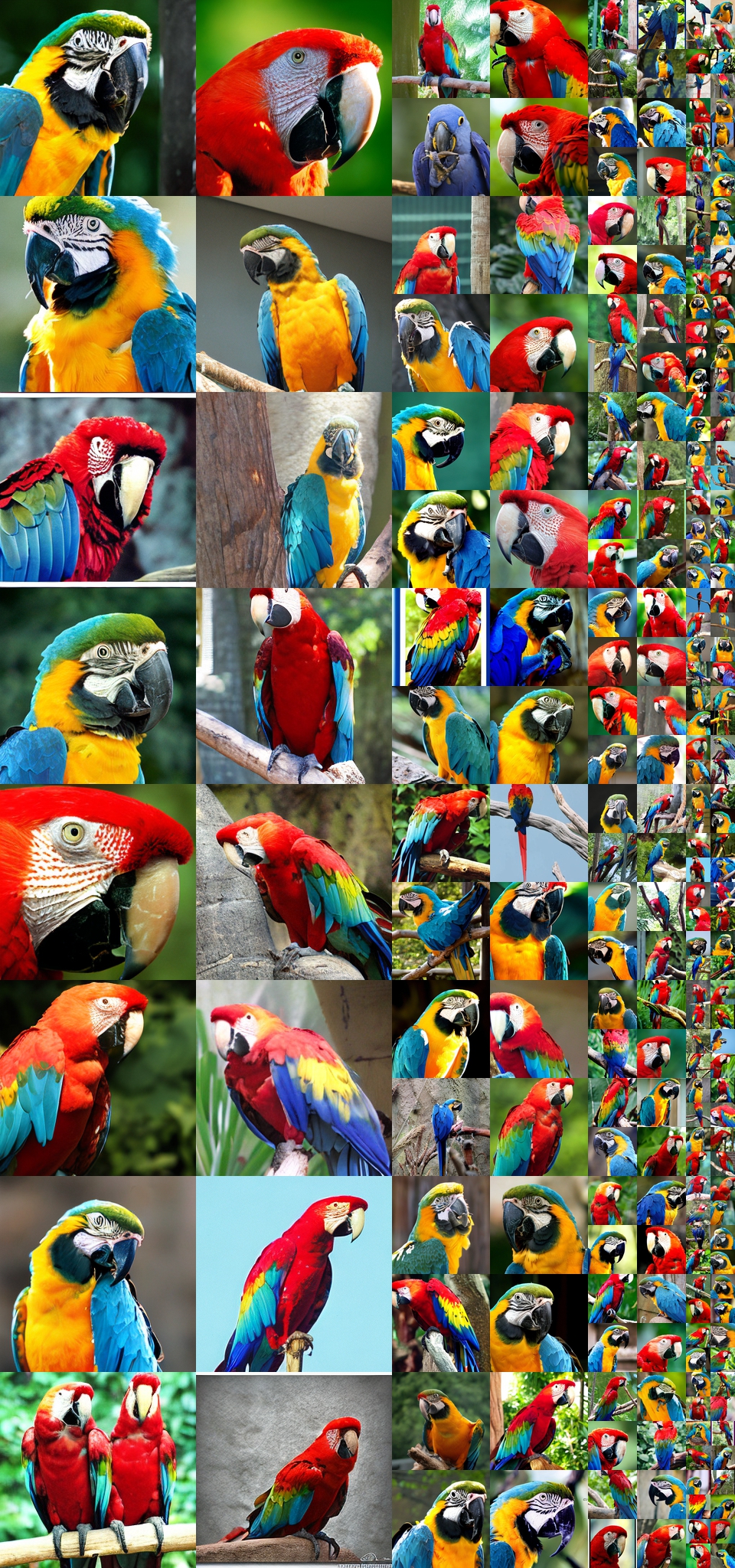}
\caption{\textbf{Uncurated $256\times256$ DiT-XL/2 samples.} \\Classifier-free guidance scale = 4.0\\Class label = ``macaw" (88)}\vspace{-2mm}
\label{fig:samples1}
\end{figure}

\begin{figure}\centering
\includegraphics[width=\linewidth]{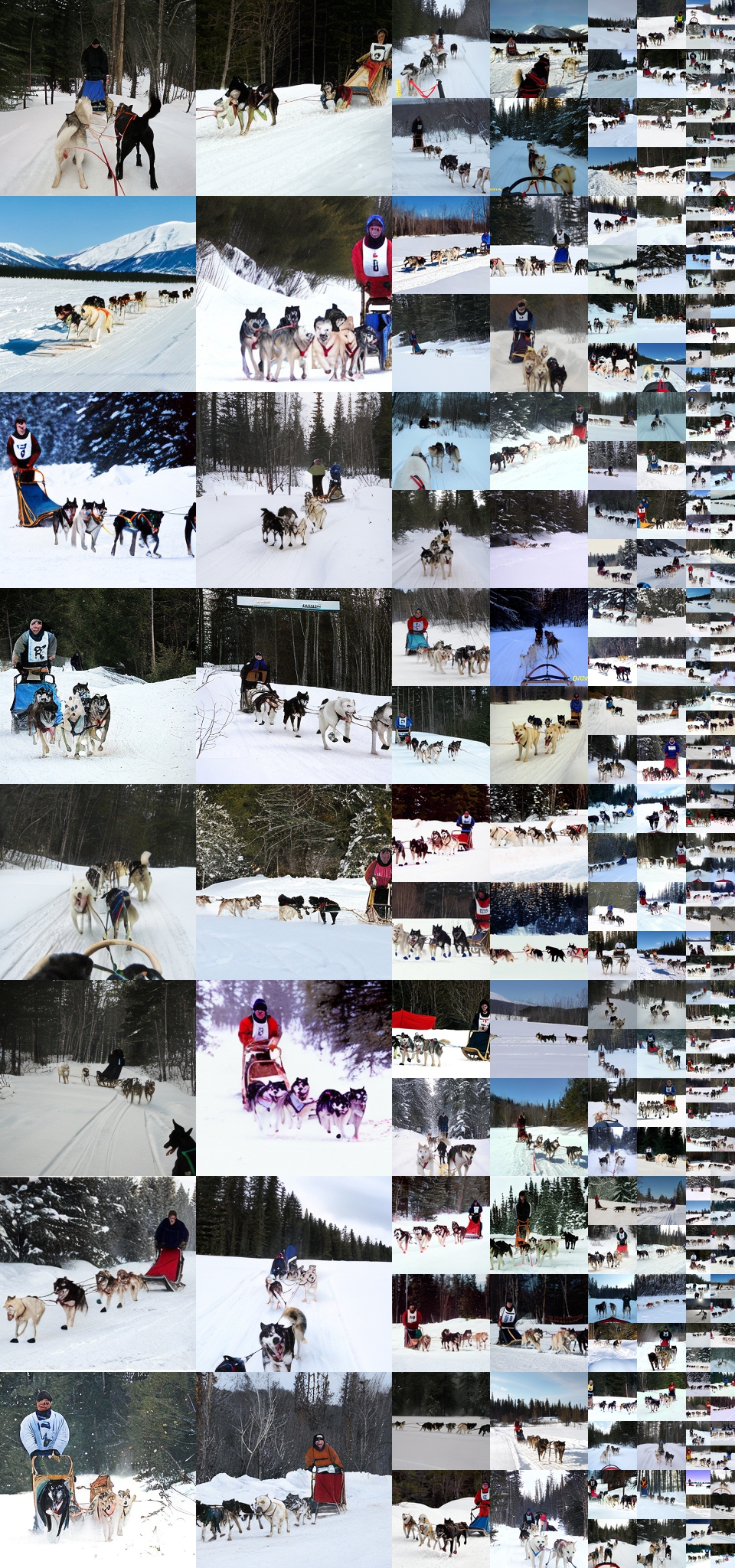}
\caption{\textbf{Uncurated $256\times256$ DiT-XL/2 samples.} \\Classifier-free guidance scale = 4.0\\Class label = ``dog sled" (537)}\vspace{-2mm}
\label{fig:samples2}
\end{figure}

\begin{figure}\centering
\includegraphics[width=\linewidth]{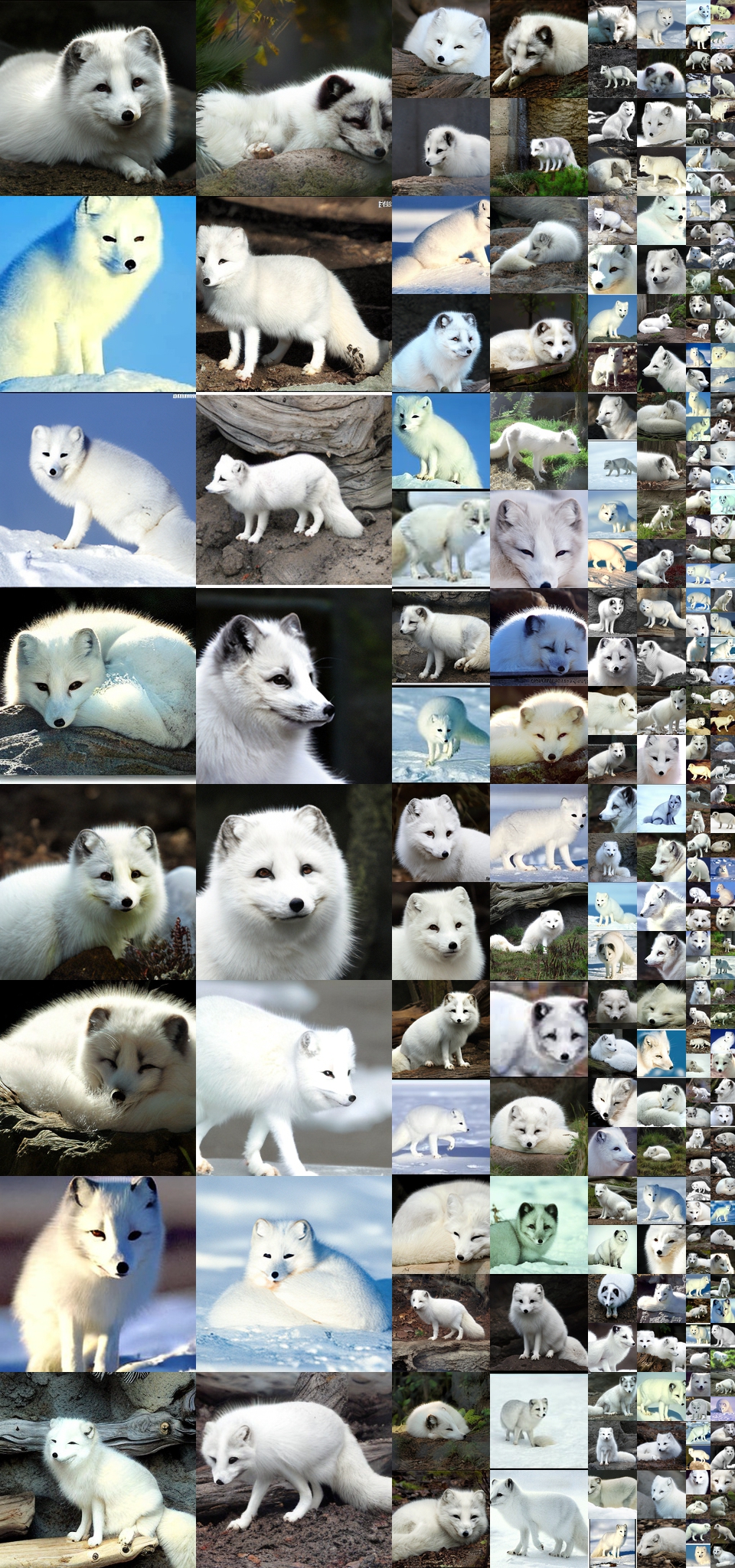}
\caption{\textbf{Uncurated $256\times256$ DiT-XL/2 samples.} \\Classifier-free guidance scale = 4.0\\Class label = ``arctic fox" (279)}\vspace{-2mm}
\label{fig:samples3}
\end{figure}

\begin{figure}\centering
\includegraphics[width=\linewidth]{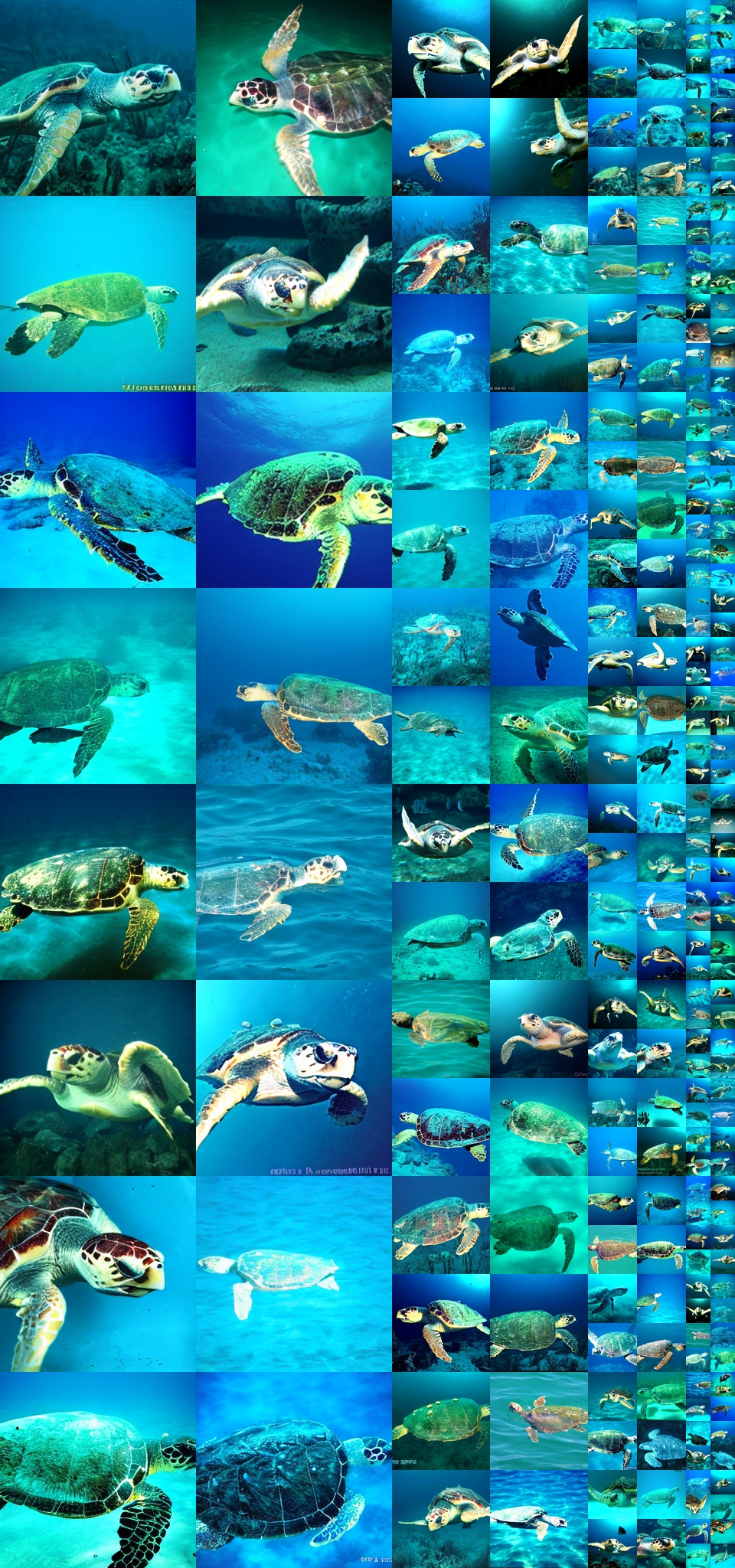}
\caption{\textbf{Uncurated $256\times256$ DiT-XL/2 samples.} \\Classifier-free guidance scale = 4.0\\Class label = ``loggerhead sea turtle" (33)}\vspace{-2mm}
\label{fig:samples4}
\end{figure}

\begin{figure}\centering
\includegraphics[width=\linewidth]{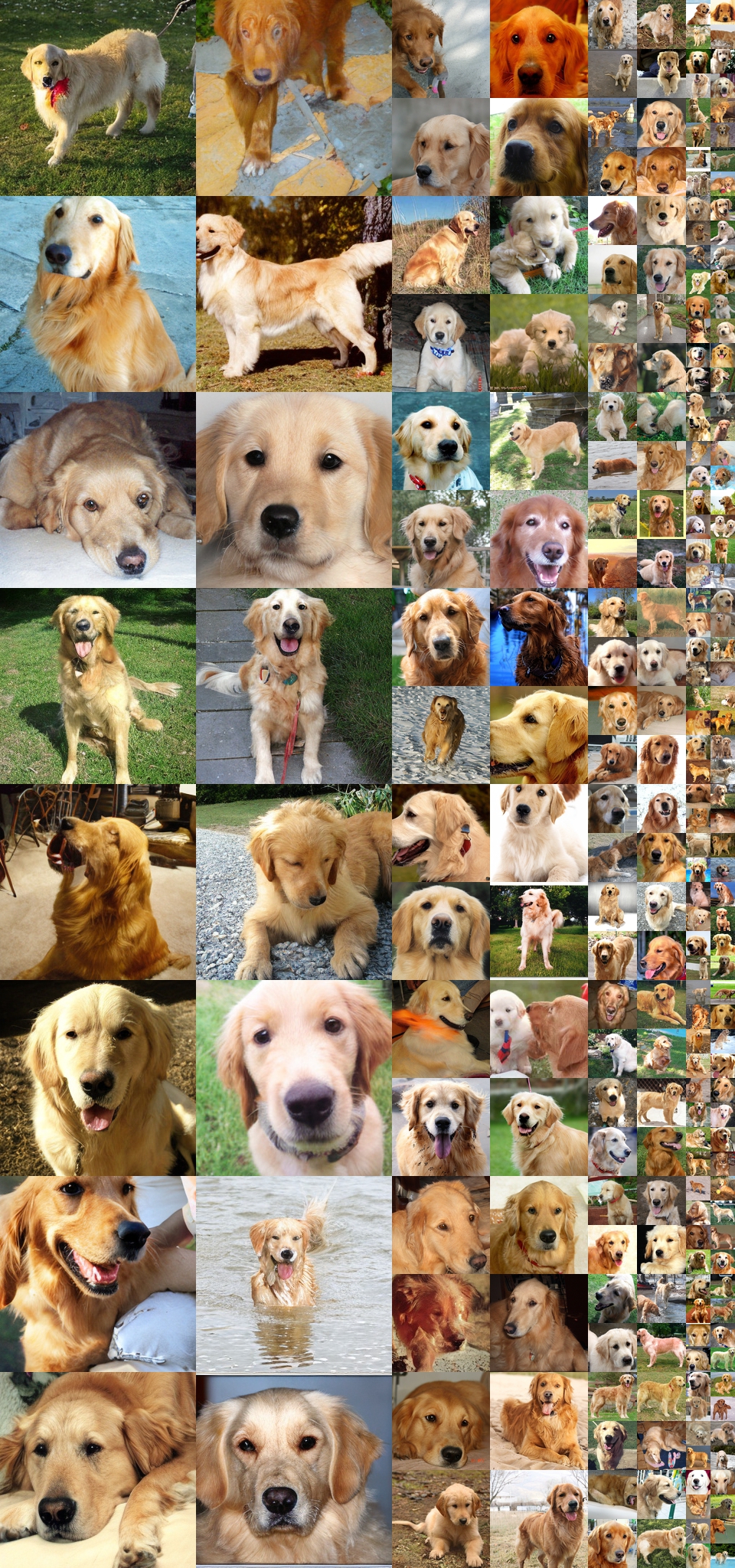}
\caption{\textbf{Uncurated $256\times256$ DiT-XL/2 samples.} \\Classifier-free guidance scale = 2.0\\Class label = ``golden retriever" (207)}\vspace{-2mm}
\label{fig:samples5}
\end{figure}

\begin{figure}\centering
\includegraphics[width=\linewidth]{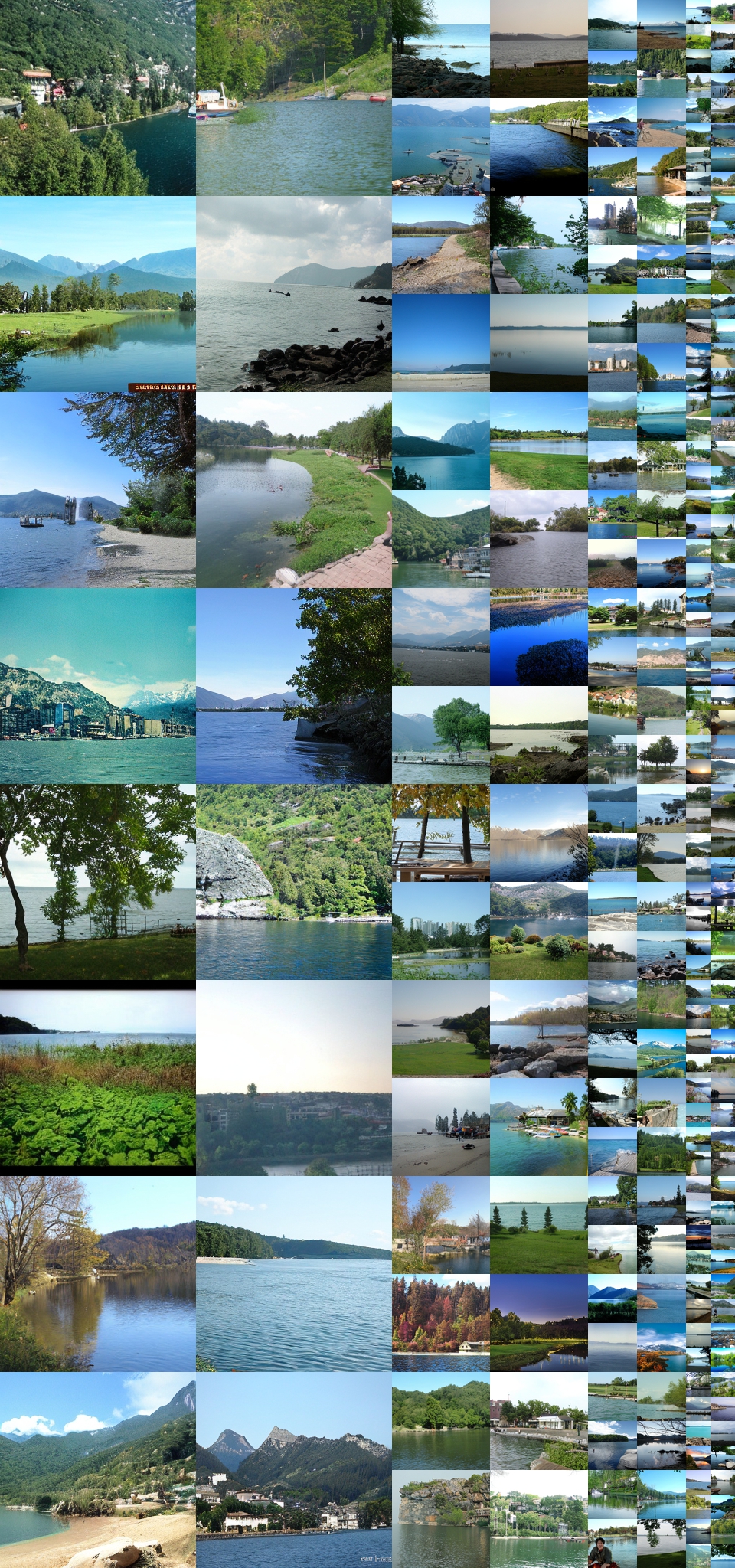}
\caption{\textbf{Uncurated $256\times256$ DiT-XL/2 samples.} \\Classifier-free guidance scale = 2.0\\Class label = ``lake shore" (975)}\vspace{-2mm}
\label{fig:samples6}
\end{figure}

\begin{figure}\centering
\includegraphics[width=\linewidth]{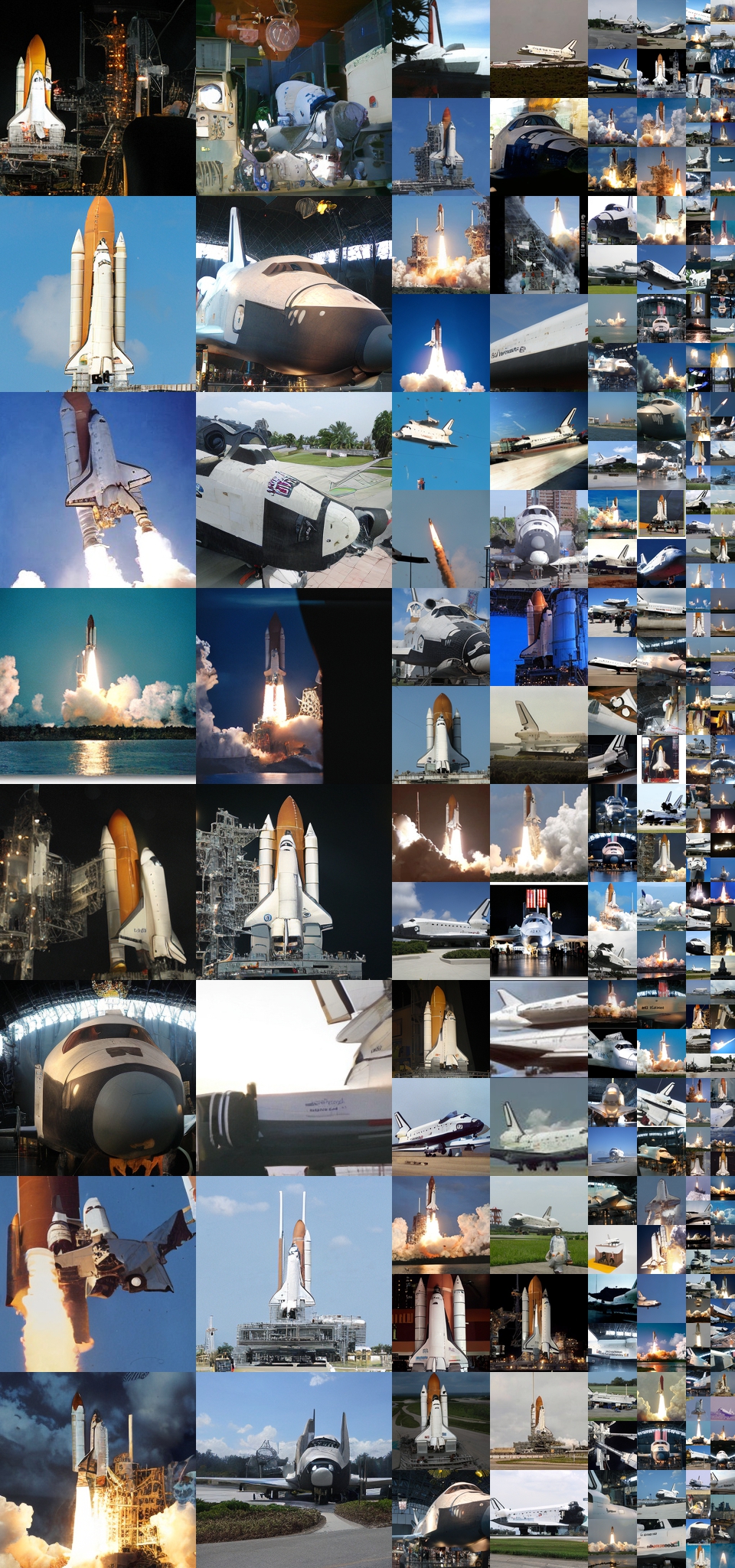}
\caption{\textbf{Uncurated $256\times256$ DiT-XL/2 samples.} \\Classifier-free guidance scale = 1.5\\Class label = ``space shuttle" (812)}\vspace{-2mm}
\label{fig:samples7}
\end{figure}

\begin{figure}\centering
\includegraphics[width=\linewidth]{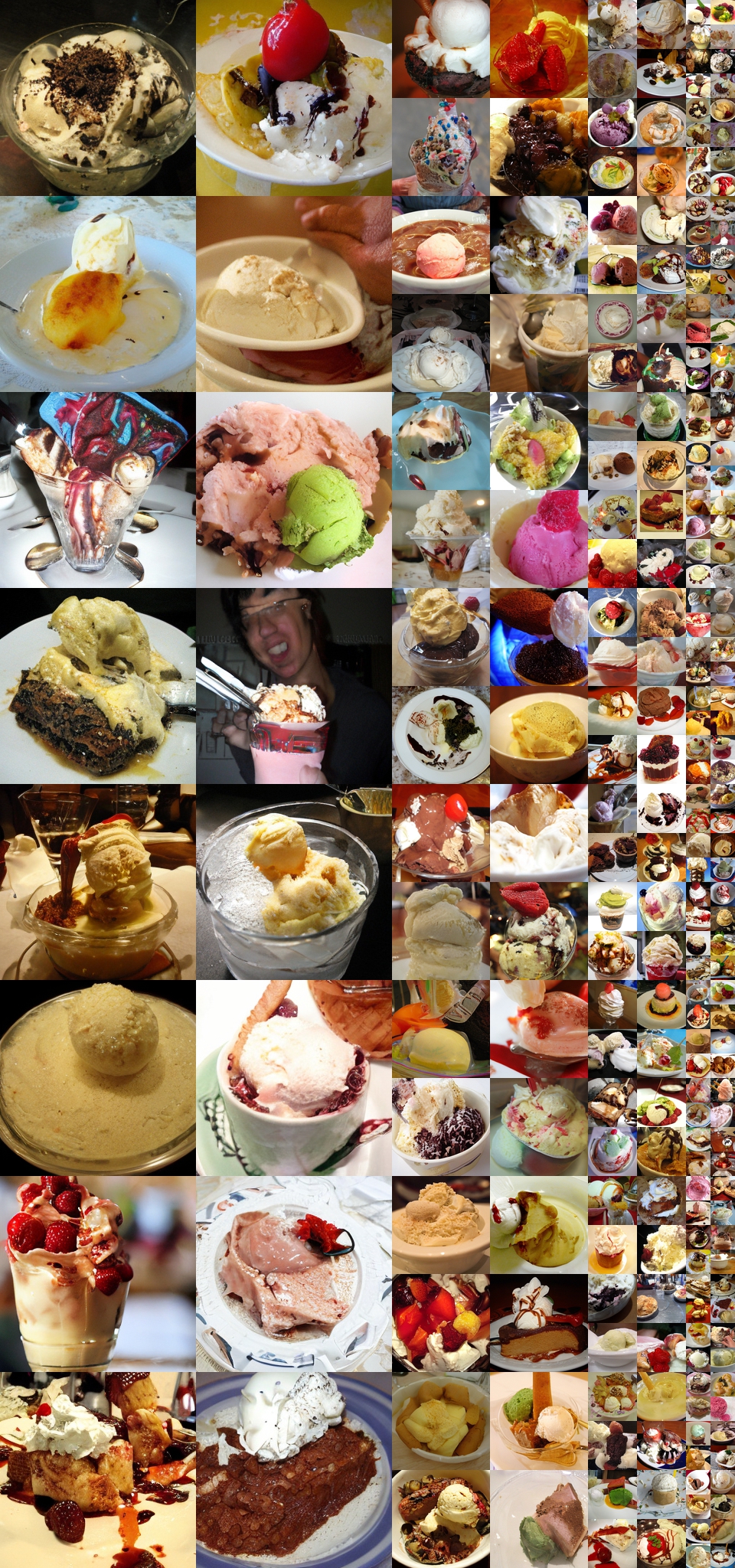}
\caption{\textbf{Uncurated $256\times256$ DiT-XL/2 samples.} \\Classifier-free guidance scale = 1.5\\Class label = ``ice cream" (928)}\vspace{-2mm}
\label{fig:samples8}
\end{figure}

We show samples from our two DiT-XL/2 models at $512\times512$ and $256\times256$ resolution trained for 3M and 7M steps, respectively. Figures~\ref{fig:teaser} and \ref{fig:sota512} show selected samples from both models. Figures~\ref{fig:samples512_1} through \ref{fig:samples8} show \textit{uncurated} samples from the two models across a range of classifier-free guidance scales and input class labels (generated with 250 DDPM sampling steps and the ft-EMA VAE decoder). As with prior work using guidance, we observe that larger scales increase visual fidelity and decrease sample diversity.

\section{Additional Scaling Results}

\paragraph{Impact of scaling on metrics beyond FID.} In Figure~\ref{fig:training_complexity_ISPR}, we show the effects of DiT scale on a suite of evaluation metrics---FID, sFID, Inception Score, Precision and Recall. We find that our FID-driven analysis in the main paper generalizes to the other metrics---across every metric, scaled-up DiT models are more compute-efficient and model Gflops are highly-correlated with performance. In particular, Inception Score and Precision benefit heavily from increased model scale.

\paragraph{Impact of scaling on training loss.} We also examine the impact of scale on training loss in Figure~\ref{fig:losses}. Increasing DiT model Gflops (via transformer size or number of input tokens) causes the training loss to decrease more rapidly and saturate at a lower value. This phenomenon is consistent with trends observed with language models, where scaled-up transformers demonstrate both improved loss curves as well as improved performance on downstream evaluation suites~\cite{kaplan2020scaling}.

\section{VAE Decoder Ablations}
We used off-the-shelf, pre-trained VAEs across our experiments. The VAE models (ft-MSE and ft-EMA) are fine-tuned versions of the original LDM ``f8" model (only the decoder weights are fine-tuned). We monitored metrics for our scaling analysis in Section~\ref{sec:experiments} using the ft-MSE decoder, and we used the ft-EMA decoder for our final metrics reported in Tables~\ref{tbl:sota} and \ref{tbl:sota512}. In this section, we ablate three different choices of the VAE decoder; the original one used by LDM and the two fine-tuned decoders used by Stable Diffusion. Because the encoders are identical across models, the decoders can be swapped-in without retraining the diffusion model. Table~\ref{tbl:decoder} shows results; XL/2 continues to outperform all prior diffusion models when using the LDM decoder.

\begin{table}[t]
    \begin{center}
    \begin{small}
    \scalebox{0.9}{
    \begin{tabular}{lccccc}
    \toprule
    \multicolumn{6}{l}{\bf{Class-Conditional ImageNet} 256$\times$256, DiT-XL/2-G (cfg=1.5)}  \\
    \toprule
    Decoder & FID$\downarrow$   & sFID$\downarrow$  & IS$\uparrow$     & Precision$\uparrow$ & Recall$\uparrow$ \\
    \toprule
    original & 2.46 & 5.18 & 271.56 & 0.82 & 0.57 \\
    ft-MSE & 2.30 & 4.73 & 276.09 & 0.83 & 0.57 \\
    ft-EMA  & 2.27 & 4.60 & 278.24 & 0.83 & 0.57 \\
    \bottomrule
    \end{tabular}
    } %
    \end{small}
    \end{center}
    \vspace{-4mm}
    \caption{\textbf{Decoder ablation.} We tested different pre-trained VAE decoder weights available at \url{https://huggingface.co/stabilityai/sd-vae-ft-mse}. Different pre-trained decoder weights yield comparable results on ImageNet $256\times256$.}
    \label{tbl:decoder}
\end{table}

\begin{table}[t]
    \begin{center}
    \begin{small}
    \scalebox{0.7}{
    \begin{tabular}{lccccc}
    \toprule
    \multicolumn{5}{l}{\textbf{Diffusion U-Net Model Complexities}}  \\
    \toprule
    Model & Image Resolution & Base Flops (G) & Upsampler Flops (G) & Total Flops (G) \\
    \toprule
    ADM   & $128\times128$ & 307 & - & 307 \\
    ADM   & $256\times256$ & 1120 & - & 1120 \\
    ADM   & $512\times512$ & 1983 & - & 1983 \\
    ADM-U & $256\times256$ & 110 & 632 & 742 \\
    ADM-U & $512\times512$ & 307 & 2506 & 2813 \\
    \toprule
    LDM-4 & $256\times256$ & 104 & - & 104 \\
    LDM-8 & $256\times256$ & 57 & - & 57 \\
    \bottomrule
    \end{tabular}
    } %
    \end{small}
    \end{center}
    \vspace{-4mm}
    \caption{\textbf{Gflop counts for baseline diffusion models that use U-Net backbones.} Note that we only count Flops for DDPM components.}
    \label{tbl:baseline-flops}
\end{table}

\clearpage
\pagestyle{fancy}
\fancyhead{}
\fancyhead[RO,LE]{\textbf{DiT-XL/2 $512\times512$ samples, classifier-free guidance scale = 4.0}}

\clearpage
\pagestyle{fancy}
\fancyhead{}
\fancyhead[RO,LE]{\textbf{DiT-XL/2 $512\times512$ samples, classifier-free guidance scale = 4.0}}

\clearpage
\pagestyle{fancy}
\fancyhead{}
\fancyhead[RO,LE]{\textbf{DiT-XL/2 $512\times512$ samples, classifier-free guidance scale = 4.0}}

\clearpage
\pagestyle{fancy}
\fancyhead{}
\fancyhead[RO,LE]{\textbf{DiT-XL/2 $512\times512$ samples, classifier-free guidance scale = 4.0}}

\clearpage
\pagestyle{fancy}
\fancyhead{}
\fancyhead[RO,LE]{\textbf{DiT-XL/2 $512\times512$ samples, classifier-free guidance scale = 2.0}}

\clearpage
\pagestyle{fancy}
\fancyhead{}
\fancyhead[RO,LE]{\textbf{DiT-XL/2 $512\times512$ samples, classifier-free guidance scale = 1.5}}

\clearpage
\pagestyle{fancy}
\fancyhead{}
\fancyhead[RO,LE]{\textbf{DiT-XL/2 $256\times256$ samples, classifier-free guidance scale = 4.0}}

\clearpage
\pagestyle{fancy}
\fancyhead{}
\fancyhead[RO,LE]{\textbf{DiT-XL/2 $256\times256$ samples, classifier-free guidance scale = 4.0}}

\clearpage
\pagestyle{fancy}
\fancyhead{}
\fancyhead[RO,LE]{\textbf{DiT-XL/2 $256\times256$ samples, classifier-free guidance scale = 2.0}}

\clearpage
\pagestyle{fancy}
\fancyhead{}
\fancyhead[RO,LE]{\textbf{DiT-XL/2 $256\times256$ samples, classifier-free guidance scale = 1.5}}

\end{document}